\documentclass[11pt]{article}
\usepackage[preprint]{acl}

\usepackage{times}
\usepackage{latexsym}
\usepackage[T1]{fontenc}
\usepackage[utf8]{inputenc}
\usepackage{microtype}
\usepackage{inconsolata}
\usepackage{graphicx}
\usepackage{amsmath,amssymb}
\usepackage{booktabs}
\usepackage{colortbl}
\usepackage{algorithm}
\usepackage{algpseudocode}
\usepackage[most]{tcolorbox}
\usepackage{tikz}
\usetikzlibrary{arrows.meta,matrix,positioning}
\usepackage{xspace}
\usepackage{enumitem}
\title{Envs-FORGE: Frontier-Optimized Reward-Grounded Environment Synthesis for Agent RL}
\author{
\bf Xiaojun Wu$^{\ast\,1,2}$,
Cehao Yang$^{\ast\,1,2}$,
Honghao Liu$^{\ast\,1,2}$,
Xueyuan Lin$^{\ast\,2}$ \\
\bf ZhiChao Shi$^{1,3}$,
Hao Zhou$^{1,3}$,
Xuhui Jiang$^{1,3}$,
Chengjin Xu$^{1,3}$ \\
\bf Jia Li$^{\dagger\,2}$,
Jian Guo$^{\dagger\,1}$
\\
\\
$^{1}$IDEA Research \\
$^{2}$The Hong Kong University of Science and Technology (Guangzhou) \\
$^{3}$DataArcTech Ltd. \\
}

\newcommand{\method}{Envs-FORGE\xspace}
\newcommand{\figref}[1]{Figure~\ref{#1}}

\begin{document}
\maketitle
{
  \renewcommand{\thefootnote}%
    {\fnsymbol{footnote}}
  \footnotetext[1]{Equal Contribution}
  \footnotetext[2]{Corresponding Author}
}

\begin{abstract}
Reinforcement learning (RL) for terminal agents needs executable training environments with reliable rewards and useful difficulty. Fixed recipes such as few-shot, Self-Instruct, and Evol-Instruct apply the same prompting policy to every seed, even when the current policy would benefit from a harder, easier, or simply different task. We present \method, a prompting policy that converts verifier rewards into per-seed environment-synthesis actions. \method{} estimates seed pass rates, scores six projection--direction actions around a target learning frontier, and solves a per-seed mixed-integer linear program (MILP) to choose the action that conditions generation. The selected action drives synchronized rewriting of the instruction, fixtures, oracle solution, tests, and Docker environment; only gold-verified bundles enter RL training. The indexed MILP form also supports optional soft skill coverage for portfolio planning. On Qwen 3.5 35B, \method{} improves Pass@1 over Base by 9.2 percentage points on tb-core (40.0\% to 49.2\%) and 6.4 points on tb-2.0 (23.0\% to 29.4\%), exceeding the strongest fixed-recipe baseline by 2.4 and 2.1 points. It reaches 77.1\% on SWE-bench Verified versus 73.4\% for Base, and improves tb-core by 6.8--9.2 points across the evaluated 4B--35B models. All synthesis methods export 100 verified environments and use 2.27M--2.88M synthesis tokens, placing the comparison at the same downstream training-set size and the same operational scale.
The source code is available at \url{https://github.com/DataArcTech/DataArc-SynData-Toolkit/}.

\end{abstract}

\section{Introduction}

Terminal agents are becoming practical interfaces for software engineering, system administration, and command-line data work. SWE-agent and OpenHands provide interactive execution loops, and Terminal-Bench and SWE-bench Verified evaluate containerized terminal tasks and repository repair \citep{yang2024sweagentagentcomputerinterfacesenable, wang2025openhandsopenplatformai, merrill2026terminalbenchbenchmarkingagentshard, swebenchverified}. RL training for these agents needs more than prompts or transcripts: it needs executable environments whose tests provide reliable rewards and whose difficulty matches the current policy.

We study the prompting policy for environment synthesis. Few-shot, Self-Instruct, Evol-Instruct, and agentic prompting flows define how an LLM generates or rewrites tasks \citep{wang2023selfinstructaligninglanguagemodels, xu2025wizardlmempoweringlargepretrained, mitra2024agentinstructgenerativeteachingagentic}. Terminal-synthesis systems add useful auxiliary components, including healthy-runtime inversion, skill-graph sampling, error injection, repository construction, procedural generation, and capability-gap discovery \citep{lin2026cligymscalableclitask, fan2026scalableterminaltasksynthesis, zhu2026termigenhighfidelityenvironmentrobust, wu2026largescaleterminalagentictrajectory, gandhi2026endlessterminalsscalingrl, dong2026agentworldscalingrealworldenvironment}. These components can be paired with many prompting policies. Our comparison isolates the policy choice itself: for a fixed synthesis and verification pipeline, which operation should the LLM apply to each seed?

\begin{figure*}[t]
  \centering
  \includegraphics[width=\textwidth]{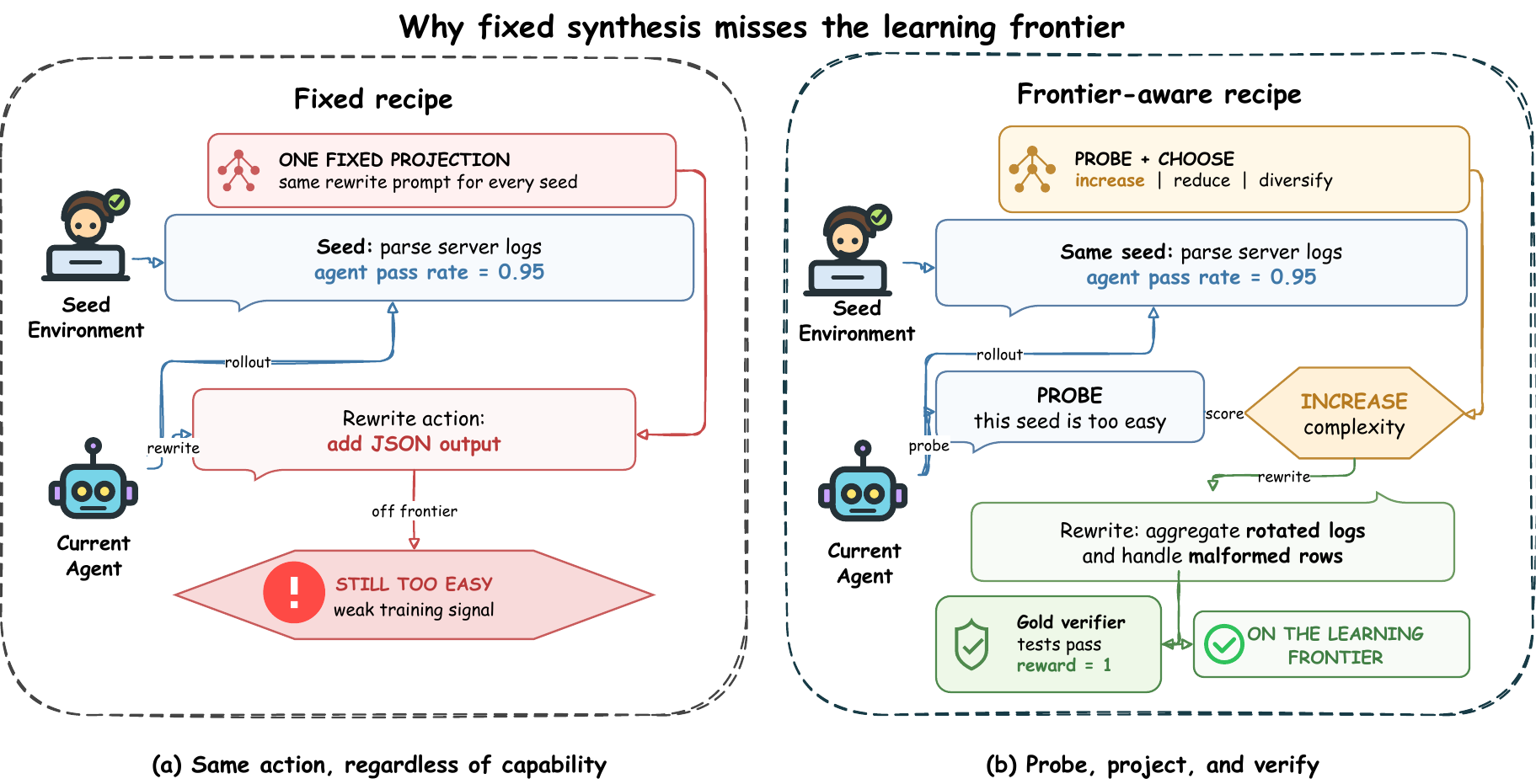}
  \caption{Motivation contrast. Fixed prompt recipes apply the same rewrite strategy to every seed, while \method{} estimates the seed's pass rate first and then chooses among increase, reduce, and diversify projections to land near the learning frontier.}
  \label{fig:motivation}
\end{figure*}

This paper asks a concrete selection question: given a seed task and a current policy, which synthesis action should place the resulting environment near the learning frontier? \figref{fig:motivation} shows the contrast. \method{} first estimates the seed pass rate from verifier rewards. It then scores six actions formed by a projection (increase, reduce, or diversify) and an evolution direction (in-depth or in-breadth). A per-seed MILP selects the action under feasibility constraints, and the chosen action conditions joint rewriting of the instruction, fixtures, oracle solution, tests, and Docker environment. The output is accepted only after gold verification.

A task's value for RL depends on its difficulty relative to the policy, not only on fluent wording. DART-Math allocates more response trials to difficult queries, and targeted tabular synthesis steers generators toward hard observations \citep{tong2024dart, ferracci2024targeted}. \method{} applies the same frontier idea to executable environment construction: the decision variable is the transformation applied to a seed. Fixed recipes become restricted masks in the same action space. Few-shot preserves seed structure, Self-Instruct invents a related task, and Evol-Instruct fixes the direction of evolution. \method{} can choose both projection and direction, including a reduce-complexity projection that creates bridge tasks from seeds beyond the current policy.

On Qwen 3.5 35B trained with Group Relative Policy Optimization (GRPO), \method{} improves Pass@1 over Base from 40.0\% to 49.2\% on tb-core and from 23.0\% to 29.4\% on tb-2.0. It exceeds the strongest fixed-recipe baseline on each benchmark by 2.4 and 2.1 percentage points, respectively. All four synthesis methods operate at a broadly comparable scale: total token use ranges from 2.27M to 2.88M while each exports exactly 100 verified environments. We report this resource accounting alongside downstream performance.

Our main contributions are:
\begin{itemize}
  \item We recast agent environment synthesis as reward-grounded action selection over verified seed tasks, with explicit increase, reduce, and diversify projections and in-depth/in-breadth evolution directions.
  \item We formulate per-task projection--direction selection as a mixed-integer program, represent the approximate semantics of fixed synthesis recipes as restricted action masks, and provide an indexed extension for optional soft skill coverage alongside synchronized artifact rewriting and gold verification.
  \item We evaluate against few-shot, Self-Instruct, and Evol-Instruct on tb-core, tb-2.0, and SWE-bench Verified, analyze tb-core across Qwen 3.5 model sizes, and report synthesis-resource accounting and qualitative environment traces.
\end{itemize}

\section{Related Work}
\label{sec:related_work}

\paragraph{Terminal agents and benchmarks.}
Terminal interaction is a central setting for agentic models. SWE-agent and OpenHands instantiate tool-using execution loops, and Terminal-Bench and SWE-bench Verified evaluate containerized terminal tasks and repository repair \citep{yang2024sweagentagentcomputerinterfacesenable, wang2025openhandsopenplatformai, merrill2026terminalbenchbenchmarkingagentshard, swebenchverified}. These benchmarks provide executable evaluation targets, not a policy-conditioned rule for synthesizing new training environments.

\paragraph{Synthetic instruction and environment generation.}
Self-Instruct, WizardLM, and AgentInstruct expand instruction data through self-generated examples, complexity evolution, or agentic flows \citep{wang2023selfinstructaligninglanguagemodels, xu2025wizardlmempoweringlargepretrained, mitra2024agentinstructgenerativeteachingagentic}. Terminal-specific systems add components that change the synthesis substrate: CLI-Gym reverts healthy Docker runtimes to failure states; SkillSynth samples skill-graph paths; TermiGen constructs containers and injects trajectory errors; TerminalTraj builds Dockerized tasks from repositories; Endless Terminals procedurally generates verified tasks; and Agent-World discovers tasks from tool ecosystems and capability gaps \citep{lin2026cligymscalableclitask, fan2026scalableterminaltasksynthesis, zhu2026termigenhighfidelityenvironmentrobust, wu2026largescaleterminalagentictrajectory, gandhi2026endlessterminalsscalingrl, dong2026agentworldscalingrealworldenvironment}. These additions are compatible with our line of work as auxiliary runtime, perturbation, construction, or discovery components. \method{} addresses a different layer: the prompting policy that chooses how an LLM should transform each seed. We compare against few-shot, Self-Instruct, and Evol-Instruct as same-level fixed policies under a shared environment-generation and verification pipeline.

\paragraph{Difficulty-aware selection.}
Our formulation is related to difficulty-aware data selection and hardness-aware synthesis. DART-Math assigns more response-generation trials to difficult queries, and targeted tabular synthesis trains generators on observations identified as hard \citep{tong2024dart, ferracci2024targeted}. Their decision concerns where to spend synthesis effort. Our decision concerns which transformation to apply to an executable seed. \method{} estimates the effect of each projection with a fixed transfer prior and selects the action closest to the learning frontier. Executable verification then checks whether the materialized bundle is runnable, gradeable, and internally consistent.

\section{Method}

\subsection{Problem Setting and Overview}
We study environment synthesis for terminal agents trained with reinforcement learning. A seed task is represented as
\(s_i=(I_i,D_i,S_i,T_i,E_i)\), where \(I_i\) is the instruction, \(D_i\) is the fixture and data bundle, \(S_i\) is an oracle solution, \(T_i\) is the test suite, and \(E_i\) is the executable environment. These components are coupled: changing the instruction without updating the solution or tests can produce a task that is either ungradeable or inconsistent with its stated objective.

\begin{figure*}[t]
  \centering
  \includegraphics[width=\textwidth]{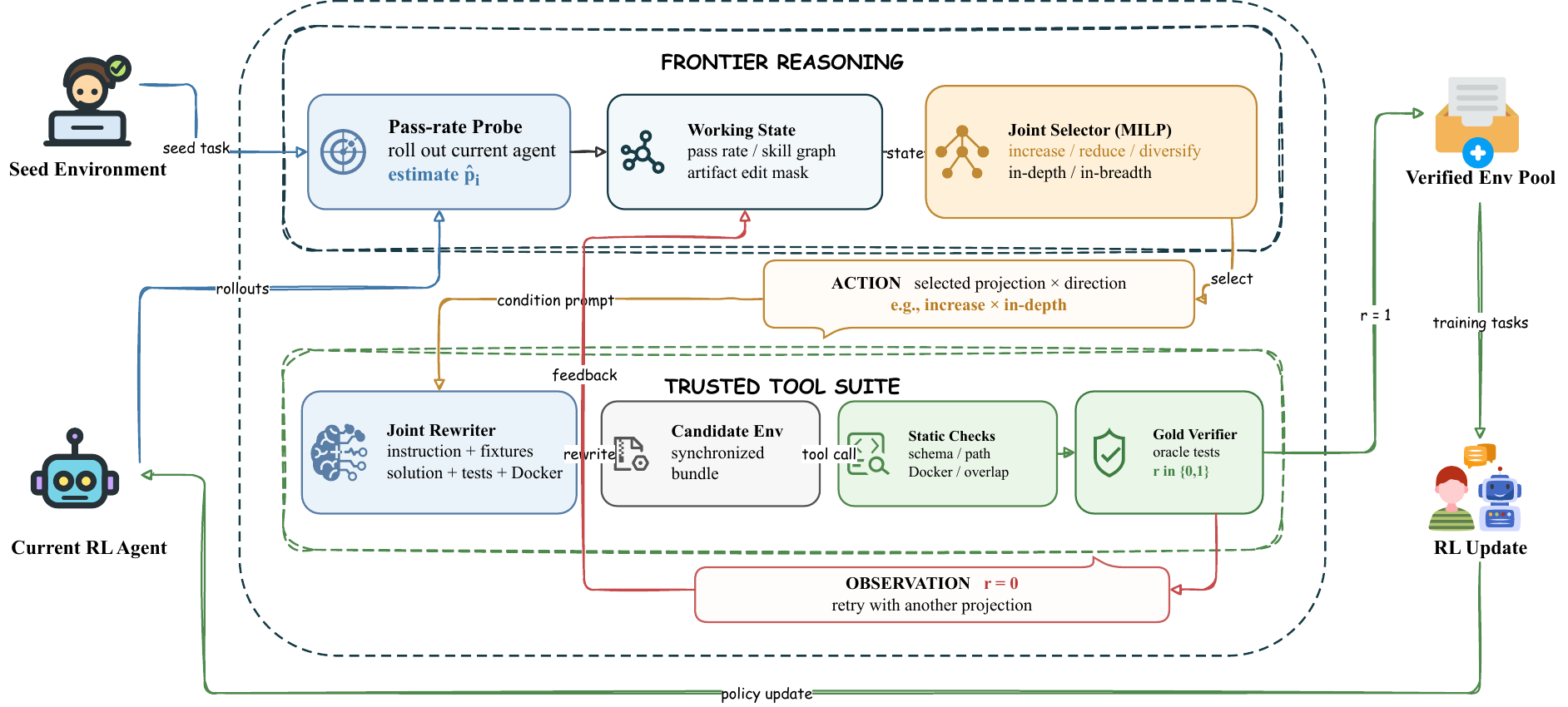}
  \caption{\method{} pipeline. Seed tasks are scored by a frontier-aware pass-rate estimator, and the MILP jointly chooses a projection and evolution direction before the synthesis model rewrites the full environment bundle. Only tasks that pass static checks and gold verification enter GRPO training.}
  \label{fig:framework}
\end{figure*}

Our method separates two decisions that fixed synthesis recipes merge into one prompt. First, it decides how a seed should move relative to the current agent. Second, it rewrites and verifies the complete environment under that decision. The projection set is
\[
\mathcal{A}=\{\texttt{increase},\texttt{reduce},\texttt{diversify}\},
\]
and the evolution-direction set is
\[
\mathcal{D}=\{\texttt{in\_depth},\texttt{in\_breadth}\}.
\]
Each candidate action is \(c_{i,a,d}=(s_i,a,d)\), with \(a\in\mathcal{A}\) and \(d\in\mathcal{D}\). Figure~\ref{fig:framework} summarizes the pipeline. \method{} estimates a frontier score for each candidate, solves one per-seed action-selection problem, conditions the synthesis prompt on the selected action, and accepts only environments that pass static checks and gold verification. The reported \method{} run uses the MILP selector described below.

\subsection{Frontier-Aware Candidate Scoring}
\label{sec:method_selection}
The selector uses verifier rewards from the current policy. For seed \(i\), let \(r_{i,t}\in[0,1]\) be the verifier reward from rollout \(t\). We estimate the seed pass rate as
\[
\hat p_i=\frac{1}{n_i}\sum_{t=1}^{n_i}r_{i,t},
\]
where binary rewards yield the empirical success rate and partial rewards, when available, contribute proportionally. The estimate is computed before candidate construction and remains fixed during optimization. We approximate the post-projection pass rate of candidate \(c_{i,a,d}\) as
\begin{equation}
\tilde p_{i,a,d} = \operatorname{clip}\!\left(\hat p_i + \Delta_a\gamma_d,\,0,\,1\right),
\label{eq:transfer}
\end{equation}
where \(\Delta_{\texttt{increase}}=-0.25\), \(\Delta_{\texttt{reduce}}=0.25\), and \(\Delta_{\texttt{diversify}}=0\) encode the intended difficulty movement. The factors \(\gamma_{\texttt{in\_depth}}=1\) and \(\gamma_{\texttt{in\_breadth}}=0.65\) attenuate breadth changes. These fixed transfer priors rank candidates before generation; they are not empirical claims that every materialized task shifts pass rate by the same amount.

We map the predicted pass rate to a learning-frontier score
\begin{equation}
\begin{aligned}
F_{i,a,d}&=\exp\!\left(-\frac{(\tilde p_{i,a,d}-\tau)^2}{2\sigma^2}\right),\\
\tau&=0.5,\qquad \sigma=0.2,
\end{aligned}
\label{eq:frontier_score}
\end{equation}
so that candidates near the target frontier receive higher scores. The three projections have distinct roles. \texttt{increase} adds concrete constraints, edge cases, stricter outputs, or larger fixtures for easy seeds. \texttt{reduce} removes secondary systems or creates a smaller bridge task for seeds beyond the current policy. \texttt{diversify} changes fixtures or neighboring requirements while preserving approximately the same difficulty and skill family.

Each candidate carries a skill-node set and a consistency contract. Let \(R_{i,a,d}\) be required skill nodes and \(O_{i,a,d}\) be optional nodes that the candidate may add or remove. Before solving, the metadata records prompt eligibility, length, split membership, and known overlap indicators. Schema, Docker, and executable-test checks are applied after materialization. This division lets the selector reason over frontier value, skill coverage, and feasibility signals while leaving semantic artifact consistency to executable verification.

\subsection{Per-Task MILP Action Selection}
\label{sec:method_milp}
The optimization unit is one seed task and its six candidate actions. For a given seed, the MILP selects one projection and one evolution direction; linked skill variables expose the subgraph induced by that action. We write the equations in indexed form over seeds so independent per-task instances can be stacked for implementation and an optional portfolio mode can add shared skill-coverage targets. We introduce binary variables
\[
x_{i,a,d}\in\{0,1\},\qquad u_{i,a,d,v}\in\{0,1\},
\]
where \(x_{i,a,d}=1\) selects action \((a,d)\) for seed \(i\), and \(u_{i,a,d,v}=1\) activates skill node \(v\) for that candidate. For active skill-coverage targets \(m_v\), we introduce continuous slack \(0\leq \xi_v\leq\bar{\xi}\); the reported experiments use \(\bar{\xi}=0.2\). Because the frontier scores and candidate metadata are precomputed before solving, the following objective is linear in the decision variables:
\begingroup
\setlength{\abovedisplayskip}{6pt}
\setlength{\abovedisplayshortskip}{6pt}
\setlength{\belowdisplayskip}{6pt}
\setlength{\belowdisplayshortskip}{6pt}
\begin{equation}
\begin{aligned}
\max_{x,u,\xi}\quad
&\sum_{i,a,d} x_{i,a,d}F_{i,a,d}
-\varepsilon\sum_{i,a,d}\sum_{v\in O_{i,a,d}}u_{i,a,d,v}\\
&-\lambda\sum_v\xi_v,
\end{aligned}
\label{eq:milp_objective}
\end{equation}
\endgroup
where \(\varepsilon=10^{-6}\) removes arbitrary activation of optional nodes and \(\lambda=0.25\) penalizes unmet coverage when coverage targets are active. The objective prioritizes frontier value, prefers compact skill realizations, and records bounded coverage shortfalls through \(\xi_v\).

The core constraints enforce local action cardinality and action--skill consistency. Here, \(N\) is the number of selected seed--action instances in the indexed form. A per-task instance selects one action; the reported stacked export has \(N=|\mathcal{S}|\). The quantities \(m_v\) and \(\xi_v\) are instantiated for active skill-coverage targets, with the slack cap \(\bar{\xi}=0.2\) in the reported experiments. Candidates that fail eligibility checks, configured overlap filters, or the prompt-length budget are removed from the feasible set; equivalently, they can be represented by \(x_{i,a,d}\leq v_{i,a,d}\), \(x_{i,a,d}\leq 1-\ell_{i,a,d}\), and \(p_{i,a,d}x_{i,a,d}\leq P_{\max}\), where \(v_{i,a,d}\), \(\ell_{i,a,d}\), and \(p_{i,a,d}\) denote eligibility, known overlap risk, and prompt length. Full artifact validity is checked after generation. The resulting linear constraints are
\begin{align}
\sum_{a,d}x_{i,a,d} &\leq 1 && \forall i, \label{eq:one_action}\\
\sum_{i,a,d}x_{i,a,d} &= N, && \label{eq:batch_size}\\
u_{i,a,d,v} &= x_{i,a,d} && \forall v\in R_{i,a,d}, \label{eq:required_skill}\\
u_{i,a,d,v} &\leq x_{i,a,d} && \forall v\in O_{i,a,d}, \label{eq:optional_skill}\\
u_{i,a,d,v} &= 0 && \forall v\notin R_{i,a,d}\cup O_{i,a,d}, \label{eq:forbidden_skill}\\
\sum_{i,a,d}u_{i,a,d,v}+\xi_v &\geq m_v && \forall v, \label{eq:soft_coverage}\\
0\leq \xi_v &\leq \bar{\xi} && \forall v. \label{eq:slack_cap}
\end{align}

The formulation also makes the relationship to heuristic baselines explicit. Let \(\rho_{b,a,d}\in\{0,1\}\) be a mask for baseline \(b\). Adding
\begin{equation}
x_{i,a,d}\leq \rho_{b,a,d}\qquad \forall i,a,d
\label{eq:baseline_mask}
\end{equation}
restricts the optimizer to a baseline-specific semantic view. Few-shot is approximated by \((\texttt{diversify},\texttt{in\_depth})\) because it preserves seed structure while changing concrete requirements; self-instruct is approximated by \((\texttt{diversify},\texttt{in\_breadth})\) because it invents a related task in the same domain; evol-depth fixes \(d=\texttt{in\_depth}\); and evol-breadth fixes \(d=\texttt{in\_breadth}\). These masks place fixed prompting policies in a shared action space. Envs-FORGE chooses both factors and adds the \texttt{reduce} projection for bridge-task construction.

\paragraph{Implementation boundary.}
The reported experiment instantiates the selector in its intended per-seed mode. Each of the 100 seeds solves a six-action MILP and selects one action; the trace serializes these decisions in the indexed form with \(N=100=|\mathcal{S}|\) and active coverage slack capped at \(\bar{\xi}=0.2\). Shared coverage targets and smaller portfolio sizes remain optional modes for curricula with explicit skill quotas, and Appendix~\ref{sec:app_milp_solution} reports the solver trace. The main evaluation measures the complete per-seed selector, synchronized materialization, and verification pipeline under the same 100-environment export size as the fixed-policy baselines.

\subsection{Verified Environment Synthesis}
\label{sec:method_synthesis}
After selection, a synthesis model receives the projection, evolution direction, preserved skill subgraph, and artifact-consistency contract. It rewrites the instruction, fixture bundle, oracle solution, tests, and Docker environment jointly. The prompt forbids instruction-only edits and hidden test requirements, and it requires the verifier output to be materialized in the expected reward file. The high-level action becomes a complete Terminal-Bench-style task.

We then apply schema, path-safety, length, Docker, test, and overlap checks. An environment enters the training pool only if its oracle solution obtains reward 1 under the generated tests. Gold verification is part of the data definition: generation can explore many candidates, but training uses only synchronized, runnable, and gradeable environments.

Algorithm~\ref{alg:forge_synthesis} summarizes the complete selection, joint materialization, and verification procedure.

\begin{algorithm}[t]
  \caption{Frontier-aware environment synthesis}
  \label{alg:forge_synthesis}
  \begin{algorithmic}[1]
    \Require Seed tasks $\mathcal{S}$ and current policy $\pi$
    \Ensure Verified environment batch $\mathcal{B}$
    \State $\mathcal{B}\gets\varnothing$
    \For{each seed $s_i\in\mathcal{S}$}
      \State Estimate $\hat p_i$ from policy rollouts
      \For{each $(a,d)\in\mathcal{A}\times\mathcal{D}$}
        \State Derive $R_{i,a,d}$, $O_{i,a,d}$, and the artifact contract
        \State Compute $\tilde p_{i,a,d}$ and $F_{i,a,d}$
        \State Remove candidates that fail pre-solve eligibility checks
      \EndFor
      \State Solve the six-action per-task MILP for seed $s_i$ and select $(a_i,d_i)$
      \State Jointly materialize environment bundle $\tilde s_{i,a_i,d_i}$
      \State Check its schema, paths, prompt length, overlap, and container isolation
      \State Build $\tilde s_{i,a_i,d_i}$; run its oracle solution and generated tests
      \If{the verifier returns reward $1$}
        \State $\mathcal{B}\gets\mathcal{B}\cup\{\tilde s_{i,a_i,d_i}\}$
      \EndIf
    \EndFor
    \State \Return $\mathcal{B}$
  \end{algorithmic}
\end{algorithm}

\noindent\textit{Implementation note.} The reported run applies one local MILP formulation per seed; for audit, the 100 local decisions are serialized in the indexed model in Eqs.~\ref{eq:milp_objective}--\ref{eq:slack_cap}. Active coverage slack is capped at \(\bar{\xi}=0.2\). A portfolio deployment may instead use smaller export targets or tighter coverage slack.

\subsection{GRPO Training and Evaluation}
\label{sec:method_training}
Accepted environments are converted to a common Terminal-Bench task format and used for GRPO training~\citep{grpo}. The main comparison trains Qwen 3.5 35B~\citep{yang2025qwen3technicalreport}; the model-size analysis additionally evaluates 4B, 9B, and 27B settings on tb-core. vLLM generates rollouts, and task tests provide rewards without a separate learned reward model. The training and evaluation protocols are fixed across synthesis sources within each comparison. Synthesis-resource use is reported separately and remains within the same overall scale across methods.

\section{Experiments}

\subsection{Setup}
We organize the evaluation around two questions: (1) does frontier-aware environment synthesis improve downstream agent performance over fixed prompting recipes, and (2) what synthesis work is needed to reach the shared 100-environment export size?

\paragraph{Training sources and verification.}
We compare four synthetic training sources: few-shot prompting, Self-Instruct, Evol-Instruct with both in-depth and in-breadth evolution, and \method{} with MILP selection. Each run continues generation, repair, and verification until exactly 100 environment bundles are accepted and exported as the downstream RL training source. All synthesized conditions contribute the same number of verified training tasks. Source records, materialized task directories, attempts, and tokens measure the synthesis-stage work required to reach that endpoint. The Base row uses no synthesized training data.

\paragraph{Training and evaluation.}
The main comparison uses Qwen 3.5 35B~\citep{yang2025qwen3technicalreport} trained with GRPO~\citep{grpo}; the model ablation additionally evaluates 4B, 9B, 27B, and 35B settings. Rollouts use vLLM and test-based rewards without a separate learned reward model. We evaluate Pass@1 on tb-core and tb-2.0 with a common protocol, add SWE-bench Verified~\citep{swebenchverified} in the benchmark ablation, and use tb-core for the model-size analysis. We report benchmark scores and percentage-point differences under this shared evaluation protocol.

\paragraph{Baselines and resource accounting.}
Few-shot and Self-Instruct instantiate fixed prompting recipes; Evol-Instruct combines its in-depth and in-breadth variants. These controls isolate prompting policy while holding artifact generation and verification fixed. The auxiliary components in Section~\ref{sec:related_work} can be paired with any policy, so we keep them outside this controlled comparison. \method{} uses the MILP selector in Section~\ref{sec:method_milp}. We report records, accepted outputs, task directories, attempts, and token totals. The methods materialize 194--210 task directories and use 2.27M--2.88M tokens to export 100 verified environments each; rejected records and intermediate artifacts do not enter RL training.

\begin{table*}[t]
  \caption{Training-data synthesis statistics and downstream Pass@1. Each synthesis run exports exactly 100 gold-verified environment bundles (the \emph{Accepted} column) to downstream RL, with task directories and token use remaining in the same broad range across methods. Records, task directories, attempts, and token counts are synthesis-stage totals used to reach that fixed endpoint; they do not increase the downstream training-set cardinality. The Base row uses no synthesized training data.}
  \label{tab:main_results}
  \centering
  \scriptsize
  \setlength{\tabcolsep}{2.3pt}
  \begin{tabular}{@{}llrrrrrrrcc@{}}
    \toprule
    \rowcolor{gray!30}
    \shortstack[l]{Training data\\ / model} & Method & Records & Accepted & \shortstack{Task\\dirs} & \shortstack{Attempt\\sum} & \shortstack{Prompt\\tokens} & \shortstack{Final completion\\tokens} & \shortstack{Total synthesis\\tokens} & \shortstack{tb-core\\Pass@1} & \shortstack{tb-2.0\\Pass@1} \\
    \midrule
    Base & -- & -- & -- & -- & -- & -- & -- & -- & 40.0\% & 23.0\% \\
    few-shot & \texttt{few\_shot} & 121 & 100 & 210 & 226 & 1,835,473 & 607,592 & 2,443,065 & 43.2\% & 24.1\% \\
    Self-Instruct & \texttt{self\_instruct} & 113 & 100 & 194 & 190 & 1,745,028 & 527,247 & 2,272,275 & 45.6\% & 27.3\% \\
    Evol-Instruct & \shortstack[l]{\texttt{evol\_instruct\_in\_depth} +\\\texttt{evol\_instruct\_in\_breadth}} & 112 & 100 & 195 & 207 & 1,974,723 & 543,023 & 2,517,746 & 46.8\% & 25.6\% \\
    \textbf{\method{}} & \textbf{MILP} & 120 & 100 & 203 & 291 & 1,862,146 & 618,910 & 2,881,056 & \textbf{49.2\%} & \textbf{29.4\%} \\
    \bottomrule
  \end{tabular}
\end{table*}

\subsection{Main Results}
\label{sec:main_results}
\begin{figure*}[!t]
  \centering
  \includegraphics[width=\textwidth]{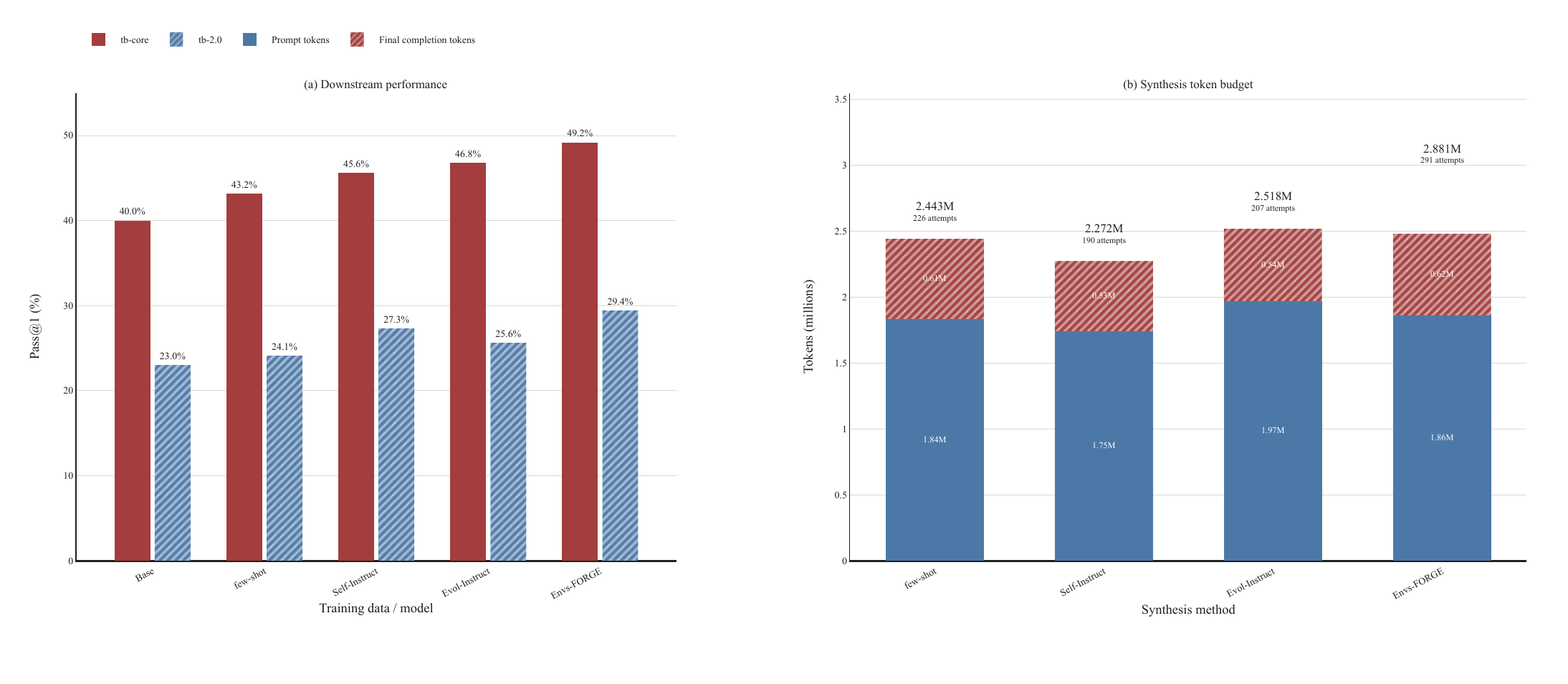}
  \caption{Downstream performance and synthesis cost. (a) Pass@1 on tb-core and tb-2.0; the Envs-FORGE run is highest on both benchmarks. (b) Synthesis-stage prompt and final-completion token totals for the four methods; labels above bars show total tokens and attempt counts. Totals occupy a comparable 2.27M--2.88M range, and each method exports exactly 100 gold-verified environments to RL. The bars measure the cost of reaching a fixed training-set size, not extra downstream training examples.}
  \label{fig:results_cost}
\end{figure*}

\paragraph{\method{} obtains the highest Pass@1 on both benchmarks.}
Table~\ref{tab:main_results} shows that the Base model obtains 40.0\% on tb-core and 23.0\% on tb-2.0. Few-shot improves these scores to 43.2\% and 24.1\%, Self-Instruct to 45.6\% and 27.3\%, and Evol-Instruct to 46.8\% and 25.6\%. \method{} reaches 49.2\% on tb-core and 29.4\% on tb-2.0, giving gains of 9.2 and 6.4 percentage points over Base. Against the strongest fixed-recipe baseline for each benchmark, the margins are 2.4 points over Evol-Instruct on tb-core and 2.1 points over Self-Instruct on tb-2.0. Figure~\ref{fig:results_cost}(a) visualizes these scores.

\paragraph{Interpretation.}
Across these runs, frontier-aware selection with verified environment synthesis yields the highest Pass@1 on both benchmarks. The fixed policies help over Base, but their single rewrite direction leaves some seeds too easy, too hard, or too close to the original task. \method{} instead chooses the seed-level action before generation.

\subsection{Ablation Studies}
\label{sec:ablations}
We next test whether the observed advantage is specific to the two Terminal-Bench-style scores or to the largest model setting.

\begin{figure*}[!t]
  \centering
  \includegraphics[width=\textwidth]{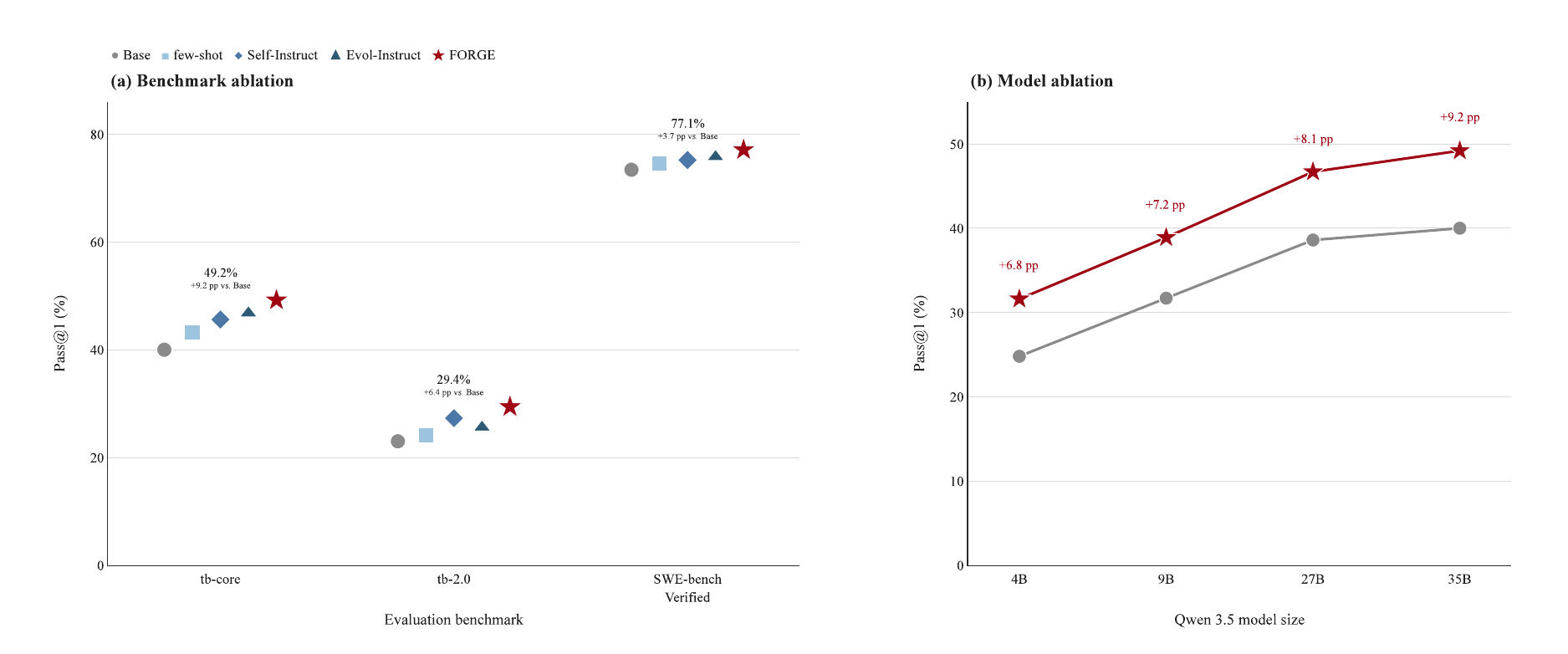}
  \caption{Benchmark and model ablations. (a) Pass@1 for the five training conditions across tb-core, tb-2.0, and SWE-bench Verified; labels show FORGE values and gains over Base. Marker shape redundantly encodes the training condition. (b) Base and FORGE tb-core Pass@1 across Qwen 3.5 model sizes; labels show the paired gain at each size.}
  \label{fig:ablation_results}
\end{figure*}

\begin{table*}[t]
  \caption{Benchmark and model ablations. In (a), bold marks the best training condition for each benchmark. Panel (b) reports tb-core, where $\Delta$ is FORGE minus Base in percentage points.}
  \label{tab:ablation_summary}
  \centering
  \small
  \begin{minipage}[t]{0.62\textwidth}
    \centering
    \textbf{(a) Benchmark ablation}\par\smallskip
    \setlength{\tabcolsep}{3.2pt}
    \begin{tabular}{@{}lrrrrr@{}}
      \toprule
      \rowcolor{gray!30}
      Benchmark & Base & few-shot & Self-Inst. & Evol-Inst. & FORGE \\
      \midrule
      tb-core & 40.0\% & 43.2\% & 45.6\% & 46.8\% & \textbf{49.2\%} \\
      tb-2.0 & 23.0\% & 24.1\% & 27.3\% & 25.6\% & \textbf{29.4\%} \\
      SWE-bench Verified & 73.4\% & 74.6\% & 75.2\% & 75.8\% & \textbf{77.1\%} \\
      \bottomrule
    \end{tabular}
  \end{minipage}\hfill
  \begin{minipage}[t]{0.34\textwidth}
    \centering
    \textbf{(b) Model ablation}\par\smallskip
    \setlength{\tabcolsep}{3pt}
    \begin{tabular}{@{}lrrr@{}}
      \toprule
      \rowcolor{gray!30}
      Model & Base & FORGE & $\Delta$ \\
      \midrule
      Qwen 3.5 4B & 24.8\% & \textbf{31.6\%} & +6.8 \\
      Qwen 3.5 9B & 31.7\% & \textbf{38.9\%} & +7.2 \\
      Qwen 3.5 27B & 38.6\% & \textbf{46.7\%} & +8.1 \\
      Qwen 3.5 35B & 40.0\% & \textbf{49.2\%} & +9.2 \\
      \bottomrule
    \end{tabular}
  \end{minipage}
\end{table*}

\paragraph{Benchmark coverage.}
The benchmark ablation adds SWE-bench Verified to tb-core and tb-2.0. As summarized in Table~\ref{tab:ablation_summary}(a) and Figure~\ref{fig:ablation_results}(a), \method{} is the highest-scoring condition on all three benchmarks: 49.2\% on tb-core, 29.4\% on tb-2.0, and 77.1\% on SWE-bench Verified. Relative to Base, the gains are +9.2, +6.4, and +3.7 percentage points; relative to the strongest fixed-recipe baseline on each benchmark, the margins are +2.4, +2.1, and +1.3 points.

\paragraph{Model coverage.}
The tb-core model ablation in Table~\ref{tab:ablation_summary}(b) and Figure~\ref{fig:ablation_results}(b) reports gains across all four Qwen 3.5 sizes. \method{} improves over Base by 6.8 points at 4B, 7.2 at 9B, 8.1 at 27B, and 9.2 at 35B.

\subsection{Synthesis Cost Analysis}
\label{sec:synthesis_cost}

\begin{table}[tb]
  \caption{Derived synthesis-cost statistics. Each synthesis run accepts 100 outputs, and all normalized costs remain within the same overall scale. Values are computed from the aggregate counts in Table~\ref{tab:main_results}.}
  \label{tab:cost_normalized}
  \centering
  \small
  \setlength{\tabcolsep}{3.5pt}
  \begin{tabular}{lrrr}
    \toprule
    \rowcolor{gray!30}
    Training data & \shortstack{Attempts /\\accepted} & \shortstack{Tokens /\\accepted} & \shortstack{Tokens /\\attempt} \\
    \midrule
    few-shot & 2.26 & 24,431 & 10,810 \\
    Self-Instruct & 1.90 & 22,723 & 11,959 \\
    Evol-Instruct & 2.07 & 25,177 & 12,163 \\
    \method{} & 2.91 & 28,811 & 9,901 \\
    \bottomrule
  \end{tabular}
\end{table}

\paragraph{Synthesis costs remain broadly comparable.}
Panel (b) of Figure~\ref{fig:results_cost} shows total synthesis use of 2.272M--2.881M tokens across the four methods, and Table~\ref{tab:main_results} shows 194--210 materialized task directories. Table~\ref{tab:cost_normalized} places tokens per accepted environment in a 22,723--28,811 range and attempts per accepted environment in a 1.90--2.91 range. These aggregates are the same order of magnitude, and each method exports exactly 100 verified environments to RL. The execution profiles differ inside this shared scale: \method{} uses 291 shorter attempts and has the lowest average tokens per attempt (9,901), while the fixed recipes use 190--226 attempts averaging 10,810--12,163 tokens. Attempts and intermediate directories characterize the synthesis route to the fixed acceptance endpoint; they do not enlarge the downstream training set.

\subsection{Evaluation Scope}
\label{sec:limitations}
The evaluation compares complete prompting policies at a fixed 100-task export and a comparable synthesis scale. Expanded component studies can isolate portfolio mode, solver-off selection, transfer-prior sensitivity, and Evol-Instruct's depth and breadth variants.

\section{Conclusion}

We present \method{}, a frontier-aware prompting policy for terminal-agent environment synthesis. The core idea is to choose a per-seed synthesis action from verifier-derived difficulty estimates, then use that action to condition synchronized and gold-verified environment rewriting. On Qwen 3.5 35B, \method{} obtains the highest Pass@1 among the evaluated methods on tb-core, tb-2.0, and SWE-bench Verified, improving over Base by 9.2, 6.4, and 3.7 percentage points. The results support a simple lesson: environment synthesis for agent RL should decide how each seed should move relative to the current policy before asking an LLM to rewrite it.

\clearpage
\section*{Limitations}
This paper evaluates the per-seed MILP mode of \method{}, which is the setting used for the reported environment-synthesis run with active coverage slack capped at \(\bar{\xi}=0.2\). The indexed formulation also supports portfolio-level skill quotas with different export targets or slack budgets, but those settings are outside the present comparison. The study compares complete prompting policies at a fixed 100-environment export size; finer component studies can test solver-off selection, transfer-prior sensitivity, and the depth and breadth variants of Evol-Instruct separately. Scaling the seed pool and model family is the natural next step.

\section*{Ethical Considerations}
This work studies environment synthesis for terminal agents. The experiments use benchmark tasks and existing task and experiment records; no human-subject data are collected. More effective environment synthesis can reduce repeated operational failures and improve the reliability of software-engineering agents, but it can also increase agent capability and persistence in command-line settings. \method{} records solver decisions and verification outcomes, applies overlap and artifact-consistency checks, and admits an environment only after gold verification so that synthesis decisions remain inspectable. These safeguards do not remove risks inherited from the base model, execution system, tools, or benchmark data. Deployment should include task-appropriate permissions, sandboxing, logging, and human oversight.

\section*{Information About Use of AI Assistants}
In preparing this manuscript, the authors used AI-assisted tools, including large language models such as GPT-5 and DeepSeek-V4, for text refinement. Their use was limited to proofreading, grammatical correction, and polishing linguistic expressions to improve clarity and readability. The authors are responsible for the final content, technical claims, citations, experimental results, and verification.

\clearpage
\bibliography{main}

@misc{gandhi2026endlessterminalsscalingrl,
      title={Endless Terminals: Scaling RL Environments for Terminal Agents}, 
      author={Kanishk Gandhi and Shivam Garg and Noah D. Goodman and Dimitris Papailiopoulos},
      year={2026},
      eprint={2601.16443},
      archivePrefix={arXiv},
      primaryClass={cs.LG},
      url={https://arxiv.org/abs/2601.16443}, 
}

@misc{zhu2026termigenhighfidelityenvironmentrobust,
      title={TermiGen: High-Fidelity Environment and Robust Trajectory Synthesis for Terminal Agents},
      author={Kaijie Zhu and Yuzhou Nie and Yijiang Li and Yiming Huang and Jialian Wu and Jiang Liu and Ximeng Sun and Zhenfei Yin and Lun Wang and Zicheng Liu and Emad Barsoum and William Yang Wang and Wenbo Guo},
      year={2026},
      eprint={2602.07274},
      archivePrefix={arXiv},
      primaryClass={cs.AI},
      url={https://arxiv.org/abs/2602.07274},
}

@misc{dong2026agentworldscalingrealworldenvironment,
      title={Agent-World: Scaling Real-World Environment Synthesis for Evolving General Agent Intelligence},
      author={Guanting Dong and Junting Lu and Junjie Huang and Wanjun Zhong and Longxiang Liu and Shijue Huang and Zhenyu Li and Yang Zhao and Xiaoshuai Song and Xiaoxi Li and Jiajie Jin and Yutao Zhu and Hanbin Wang and Fangyu Lei and Qinyu Luo and Mingyang Chen and Zehui Chen and Jiazhan Feng and Ji-Rong Wen and Zhicheng Dou},
      year={2026},
      eprint={2604.18292},
      archivePrefix={arXiv},
      primaryClass={cs.AI},
      url={https://arxiv.org/abs/2604.18292},
}

@misc{fan2026scalableterminaltasksynthesis,
      title={Toward Scalable Terminal Task Synthesis via Skill Graphs}, 
      author={Zhiyuan Fan and Tinghao Yu and Yuanjun Cai and Jiangtao Guan and Yun Yang and Dingxin Hu and Jiang Zhou and Xing Wu and Zhuo Han and Feng Zhang and Lilin Wang},
      year={2026},
      eprint={2604.25727},
      archivePrefix={arXiv},
      primaryClass={cs.AI},
      url={https://arxiv.org/abs/2604.25727}, 
}

@misc{merrill2026terminalbenchbenchmarkingagentshard,
      title={Terminal-Bench: Benchmarking Agents on Hard, Realistic Tasks in Command Line Interfaces}, 
      author={Mike A. Merrill and Alexander G. Shaw and Nicholas Carlini and Boxuan Li and Harsh Raj and Ivan Bercovich and Lin Shi and Jeong Yeon Shin and Thomas Walshe and E. Kelly Buchanan and Junhong Shen and Guanghao Ye and Haowei Lin and Jason Poulos and Maoyu Wang and Marianna Nezhurina and Jenia Jitsev and Di Lu and Orfeas Menis Mastromichalakis and Zhiwei Xu and Zizhao Chen and Yue Liu and Robert Zhang and Leon Liangyu Chen and Anurag Kashyap and Jan-Lucas Uslu and Jeffrey Li and Jianbo Wu and Minghao Yan and Song Bian and Vedang Sharma and Ke Sun and Steven Dillmann and Akshay Anand and Andrew Lanpouthakoun and Bardia Koopah and Changran Hu and Etash Guha and Gabriel H. S. Dreiman and Jiacheng Zhu and Karl Krauth and Li Zhong and Niklas Muennighoff and Robert Amanfu and Shangyin Tan and Shreyas Pimpalgaonkar and Tushar Aggarwal and Xiangning Lin and Xin Lan and Xuandong Zhao and Yiqing Liang and Yuanli Wang and Zilong Wang and Changzhi Zhou and David Heineman and Hange Liu and Harsh Trivedi and John Yang and Junhong Lin and Manish Shetty and Michael Yang and Nabil Omi and Negin Raoof and Shanda Li and Terry Yue Zhuo and Wuwei Lin and Yiwei Dai and Yuxin Wang and Wenhao Chai and Shang Zhou and Dariush Wahdany and Ziyu She and Jiaming Hu and Zhikang Dong and Yuxuan Zhu and Sasha Cui and Ahson Saiyed and Arinbjörn Kolbeinsson and Jesse Hu and Christopher Michael Rytting and Ryan Marten and Yixin Wang and Alex Dimakis and Andy Konwinski and Ludwig Schmidt},
      year={2026},
      eprint={2601.11868},
      archivePrefix={arXiv},
      primaryClass={cs.SE},
      url={https://arxiv.org/abs/2601.11868}, 
}

@misc{wu2026largescaleterminalagentictrajectory,
      title={Large-Scale Terminal Agentic Trajectory Generation from Dockerized Environments}, 
      author={Siwei Wu and Yizhi Li and Yuyang Song and Wei Zhang and Yang Wang and Riza Batista-Navarro and Xian Yang and Mingjie Tang and Bryan Dai and Jian Yang and Chenghua Lin},
      year={2026},
      eprint={2602.01244},
      archivePrefix={arXiv},
      primaryClass={cs.CL},
      url={https://arxiv.org/abs/2602.01244}, 
}

@misc{wang2025openhandsopenplatformai,
      title={OpenHands: An Open Platform for AI Software Developers as Generalist Agents}, 
      author={Xingyao Wang and Boxuan Li and Yufan Song and Frank F. Xu and Xiangru Tang and Mingchen Zhuge and Jiayi Pan and Yueqi Song and Bowen Li and Jaskirat Singh and Hoang H. Tran and Fuqiang Li and Ren Ma and Mingzhang Zheng and Bill Qian and Yanjun Shao and Niklas Muennighoff and Yizhe Zhang and Binyuan Hui and Junyang Lin and Robert Brennan and Hao Peng and Heng Ji and Graham Neubig},
      year={2025},
      eprint={2407.16741},
      archivePrefix={arXiv},
      primaryClass={cs.SE},
      url={https://arxiv.org/abs/2407.16741}, 
}

@misc{wang2023selfinstructaligninglanguagemodels,
      title={Self-Instruct: Aligning Language Models with Self-Generated Instructions}, 
      author={Yizhong Wang and Yeganeh Kordi and Swaroop Mishra and Alisa Liu and Noah A. Smith and Daniel Khashabi and Hannaneh Hajishirzi},
      year={2023},
      eprint={2212.10560},
      archivePrefix={arXiv},
      primaryClass={cs.CL},
      url={https://arxiv.org/abs/2212.10560}, 
}

@misc{mitra2024agentinstructgenerativeteachingagentic,
      title={AgentInstruct: Toward Generative Teaching with Agentic Flows}, 
      author={Arindam Mitra and Luciano Del Corro and Guoqing Zheng and Shweti Mahajan and Dany Rouhana and Andres Codas and Yadong Lu and Wei-ge Chen and Olga Vrousgos and Corby Rosset and Fillipe Silva and Hamed Khanpour and Yash Lara and Ahmed Awadallah},
      year={2024},
      eprint={2407.03502},
      archivePrefix={arXiv},
      primaryClass={cs.AI},
      url={https://arxiv.org/abs/2407.03502}, 
}

@misc{xu2025wizardlmempoweringlargepretrained,
      title={WizardLM: Empowering large pre-trained language models to follow complex instructions}, 
      author={Can Xu and Qingfeng Sun and Kai Zheng and Xiubo Geng and Pu Zhao and Jiazhan Feng and Chongyang Tao and Qingwei Lin and Daxin Jiang},
      year={2025},
      eprint={2304.12244},
      archivePrefix={arXiv},
      primaryClass={cs.CL},
      url={https://arxiv.org/abs/2304.12244}, 
}

@misc{yang2024sweagentagentcomputerinterfacesenable,
      title={SWE-agent: Agent-Computer Interfaces Enable Automated Software Engineering}, 
      author={John Yang and Carlos E. Jimenez and Alexander Wettig and Kilian Lieret and Shunyu Yao and Karthik Narasimhan and Ofir Press},
      year={2024},
      eprint={2405.15793},
      archivePrefix={arXiv},
      primaryClass={cs.SE},
      url={https://arxiv.org/abs/2405.15793}, 
}

@misc{lin2026cligymscalableclitask,
      title={CLI-Gym: Scalable CLI Task Generation via Agentic Environment Inversion}, 
      author={Yusong Lin and Haiyang Wang and Shuzhe Wu and Lue Fan and Feiyang Pan and Sanyuan Zhao and Dandan Tu},
      year={2026},
      eprint={2602.10999},
      archivePrefix={arXiv},
      primaryClass={cs.AI},
      url={https://arxiv.org/abs/2602.10999}, 
}

@misc{tong2024dart,
  title={Dart-math: Difficulty-aware rejection tuning for mathematical problem-solving},
  author={Tong, Yuxuan and Zhang, Xiwen and Wang, Rui and Wu, Ruidong and He, Junxian},
  journal={Advances in Neural Information Processing Systems},
  volume={37},
  pages={7821--7846},
  year={2024}
}

@article{ferracci2024targeted,
  title={Targeted synthetic data generation for tabular data via hardness characterization},
  author={Ferracci, Tommaso and Goldmann, Leonie Tabea and Hinel, Anton and Passino, Francesco Sanna},
  journal={arXiv preprint arXiv:2410.00759},
  year={2024}
}

@article{grpo,
  title={Deepseekmath: Pushing the limits of mathematical reasoning in open language models},
  author={Shao, Zhihong and Wang, Peiyi and Zhu, Qihao and Xu, Runxin and Song, Junxiao and Bi, Xiao and Zhang, Haowei and Zhang, Mingchuan and Li, YK and Wu, Yang and others},
  journal={arXiv preprint arXiv:2402.03300},
  year={2024}
}

@misc{yang2025qwen3technicalreport,
      title={Qwen3 Technical Report},
      author={An Yang and Anfeng Li and Baosong Yang and Beichen Zhang and Binyuan Hui and Bo Zheng and Bowen Yu and Chang Gao and Chengen Huang and Chenxu Lv and Chujie Zheng and Dayiheng Liu and Fan Zhou and Fei Huang and Feng Hu and Hao Ge and Haoran Wei and Huan Lin and Jialong Tang and Jian Yang and Jianhong Tu and Jianwei Zhang and Jianxin Yang and Jiaxi Yang and Jing Zhou and Jingren Zhou and Junyang Lin and Kai Dang and Keqin Bao and Kexin Yang and Le Yu and Lianghao Deng and Mei Li and Mingfeng Xue and Mingze Li and Pei Zhang and Peng Wang and Qin Zhu and Rui Men and Ruize Gao and Shixuan Liu and Shuang Luo and Tianhao Li and Tianyi Tang and Wenbiao Yin and Xingzhang Ren and Xinyu Wang and Xinyu Zhang and Xuancheng Ren and Yang Fan and Yang Su and Yichang Zhang and Yinger Zhang and Yu Wan and Yuqiong Liu and Zekun Wang and Zeyu Cui and Zhenru Zhang and Zhipeng Zhou and Zihan Qiu},
      year={2025},
      eprint={2505.09388},
      archivePrefix={arXiv},
      primaryClass={cs.CL},
      url={https://arxiv.org/abs/2505.09388}, 
}

@misc{swebenchverified,
  author       = {{OpenAI}},
  title        = {Introducing {SWE}-bench Verified},
  year         = {2024},
  howpublished = {\url{https://openai.com/index/introducing-swe-bench-verified/}},
}

\clearpage
\appendix

\section*{Appendix Overview}
The appendix is organized as follows.
\begin{itemize}[leftmargin=1.4em,itemsep=2pt,topsep=3pt]
  \item Appendix~\ref{sec:app_reproducibility} expands the optimization and reproducibility details, including action semantics, baseline restrictions, the solver--generation boundary, artifact-consistency checks, training and evaluation settings, and five qualitative frontier cases.
  \item Appendix~\ref{sec:app_milp_solution} describes MILP model assembly, SCIP's branch-and-cut procedure, the recorded synthesis solve, solution decoding, fallback behavior, and the boundary of the solver's correctness guarantee.
  \item Appendix~\ref{sec:app_prompts} reproduces the core baseline and \method{} prompt clauses, including strategy injection, repair, the solver payload, projection instructions, artifact edit masks, the skill-subgraph contract, and original-to-synthesized task excerpts.
  \item Appendix~\ref{sec:app_broader_impact} reports the compute resources used for synthesis, training, and evaluation.
\end{itemize}

\section{Optimization and Reproducibility Details}
\label{sec:app_reproducibility}

This appendix expands the action semantics, solver boundary, environment-verification protocol, and qualitative cases summarized in the main paper.

\subsection{Action Semantics and Baseline Restrictions}
\label{sec:app_actions}

The projection and evolution direction answer different questions. The projection specifies how difficulty should move relative to the current policy: increase an easy seed, reduce a hard seed into a bridge task, or diversify a seed already near the frontier. The direction specifies whether the rewrite follows the same skill chain (\texttt{in\_depth}) or moves to a neighboring skill or task type (\texttt{in\_breadth}). Table~\ref{tab:action_space} records the operational contracts used to instantiate these choices.

\begin{table*}[t]
  \caption{Operational semantics of the Envs-FORGE action space and its restricted baseline views. The baseline correspondences describe approximate prompt semantics, not exact equivalence between generation procedures.}
  \label{tab:action_space}
  \centering
  \small
  \setlength{\tabcolsep}{4pt}
  \begin{tabular}{@{}p{0.16\textwidth}p{0.16\textwidth}p{0.29\textwidth}p{0.31\textwidth}@{}}
    \toprule
    \rowcolor{gray!30}
    Projection & Agent-relative regime & Permitted movement & Invariants \\
    \midrule
    \texttt{increase} & Seed is too easy & Add one or two constraints, edge cases, larger fixtures, or stricter deterministic outputs. & Preserve the core skill subgraph; update the solution and tests to cover every added requirement. \\
    \texttt{reduce} & Seed is too hard & Remove secondary systems or replace brittle infrastructure with a smaller bridge task. & Preserve the target skill path and its essential artifacts; simplification must not collapse the task into an unrelated problem. \\
    \texttt{diversify} & Seed is near the frontier & Change fixtures, neighboring requirements, or scenario while keeping similar difficulty. & Avoid unrelated breadth jumps; synchronize expected outputs and verifier logic with the new fixture. \\
    \bottomrule
  \end{tabular}

  \vspace{5pt}
  \begin{tabular}{@{}p{0.19\textwidth}p{0.25\textwidth}p{0.48\textwidth}@{}}
    \toprule
    \rowcolor{gray!30}
    Restricted policy & Mask in the per-task action space & Interpretation \\
    \midrule
    few-shot & $(a,d)=(\texttt{diversify},\texttt{in\_depth})$ & Approximate semantics: generate a close seed-conditioned variant while preserving the seed's core structure. \\
    Self-Instruct & $(a,d)=(\texttt{diversify},\texttt{in\_breadth})$ & Approximate semantics: invent a related task in the same domain with synchronized artifacts. \\
    Evol-Instruct depth & $d=\texttt{in\_depth}$ & Approximate semantics: fix the evolution direction to deeper constraints or reasoning steps. \\
    Evol-Instruct breadth & $d=\texttt{in\_breadth}$ & Approximate semantics: fix the evolution direction to a neighboring skill or task type. \\
    \bottomrule
  \end{tabular}
\end{table*}

The restricted policies in the lower half of Table~\ref{tab:action_space} make the baseline-inclusion claim precise. If $\mathcal{X}_{\textsc{forge}}$ denotes the feasible set defined by Eqs.~\ref{eq:one_action}--\ref{eq:soft_coverage}, then baseline $b$ induces
\begin{equation}
\mathcal{X}_b=\left\{x\in\mathcal{X}_{\textsc{forge}}:
x_{i,a,d}\leq\rho_{b,a,d}\ \forall i,a,d\right\}.
\label{eq:restricted_feasible_set}
\end{equation}
This gives a unified optimization view of the prompting policies. The underlying prompts remain distinct: few-shot is structurally narrower than generic diversification, Self-Instruct permits freer same-domain invention, and the two Evol-Instruct variants fix an evolution direction. Envs-FORGE selects both the projection and direction for each seed from the full feasible region using policy-relative frontier scores; optional portfolio mode adds shared skill-coverage targets across those local instances.

\subsection{Candidate Construction and Solver Boundary}
\label{sec:app_solver_boundary}

Each seed induces six candidate action variants and one local MILP instance. Before solving, the pipeline estimates the seed pass rate, applies the fixed projection--direction transfer heuristic in Eq.~\ref{eq:transfer}, computes the frontier utility in Eq.~\ref{eq:frontier_score}, and derives required and optional skill nodes. Candidate prompt length, split membership, and known overlap or validity indicators are also treated as constants. The local MILP jointly selects the projection and evolution direction; the indexed equations additionally expose skill activation and optional portfolio coverage. Table~\ref{tab:skill_node_taxonomy} lists the skill-node vocabulary, and Table~\ref{tab:milp_symbols} summarizes the complete notation.

\begin{table*}[t]
  \caption{Skill-node taxonomy used for candidate metadata. These nodes define the required set \(R_{i,a,d}\) and optional addable or removable set \(O_{i,a,d}\) attached to each action candidate; they are metadata tags for the action contract rather than separate generation artifacts.}
  \label{tab:skill_node_taxonomy}
  \centering
  \small
  \setlength{\tabcolsep}{5pt}
  \begin{tabular}{@{}p{0.18\textwidth}p{0.74\textwidth}@{}}
    \toprule
    \rowcolor{gray!30}
    Category & Skill nodes \\
    \midrule
    Domain &
    \texttt{system\_admin}, \texttt{data\_processing}, \texttt{testing}, \texttt{security}, \texttt{ml\_ops}, \texttt{game\_reasoning} \\
    Tool &
    \texttt{bash}, \texttt{python}, \texttt{systemd}, \texttt{docker}, \texttt{sql}, \texttt{node} \\
    Artifact &
    \texttt{logs}, \texttt{csv}, \texttt{json}, \texttt{jsonl}, \texttt{source\_code}, \texttt{config} \\
    Operation &
    \texttt{merge}, \texttt{parse}, \texttt{repair}, \texttt{configure}, \texttt{generate\_tests}, \texttt{rank}, \texttt{validate} \\
    Constraint &
    \texttt{concurrency}, \texttt{scheduling}, \texttt{atomicity} \\
    FORGE-added nodes &
    \texttt{edge\_cases}, \texttt{structured\_output}, \texttt{deterministic\_sorting}, \texttt{alternate\_fixture}, \texttt{neighboring\_constraint}, \texttt{file\_level\_simulation}, \texttt{deterministic\_fixture} \\
    \bottomrule
  \end{tabular}
\end{table*}

\begin{table*}[t]
  \caption{MILP notation and implementation boundary. Candidate scores and metadata are constants when the solver is invoked.}
  \label{tab:milp_symbols}
  \centering
  \small
  \setlength{\tabcolsep}{5pt}
  \begin{tabular}{@{}p{0.11\textwidth}p{0.34\textwidth}p{0.11\textwidth}p{0.34\textwidth}@{}}
    \toprule
    \rowcolor{gray!30}
    Symbol & Meaning & Symbol & Meaning \\
    \midrule
    $s_i$ & Seed environment bundle & $\hat p_i$ & Current policy pass-rate estimate on seed $i$ \\
    $a,d$ & Projection and evolution direction & $\tilde p_{i,a,d}$ & Predicted pass rate after applying $(a,d)$ \\
    $F_{i,a,d}$ & Frontier utility of a candidate & $x_{i,a,d}$ & Binary candidate-selection variable \\
    $R_{i,a,d}$ & Required skill nodes & $O_{i,a,d}$ & Optional addable or removable skill nodes \\
    $u_{i,a,d,v}$ & Binary activation of skill node $v$ & $\xi_v$ & Continuous shortfall for target coverage $m_v$; capped by \(\bar{\xi}=0.2\) in reported experiments \\
    $N$ & Selected seed--action count in indexed portfolio form; one per local task & $\rho_{b,a,d}$ & Constant action mask for restricted baseline $b$ \\
    \bottomrule
  \end{tabular}
\end{table*}

The artifact edit mask is outside the solver. It is deterministically derived from the selected action and passed to generation as a consistency contract. Joint materialization and executable verification enforce semantic dependencies among instructions, fixtures, solutions, tests, and containers; the MILP handles the discrete action choice. Likewise, $\rho_{b,a,d}$ is an analysis-time constant for defining restricted policies. It is not a free variable in the default Envs-FORGE solve. Failed verification attempts are excluded from the training set; generation or repair can continue only within the synthesis budget for that run.

Section~\ref{sec:app_milp_solution} details model assembly, branch-and-cut solution, and solver-trace validation.
Section~\ref{sec:app_prompts} reproduces the core prompt clauses that connect the solver decision to generation, including the projection instruction, artifact edit mask, and skill-subgraph contract. It also records the shared baseline template and the strategy instruction that distinguishes each fixed recipe.

\subsection{Artifact-Consistency Contract}
\label{sec:app_artifact_contract}

A materialized sample is a complete executable environment, not an instruction-only rewrite. The contract requires synchronized versions of (i) the natural-language instruction, (ii) input fixtures and data, (iii) the oracle solution, (iv) executable tests and reward logic, and (v) the container environment. Added requirements must be visible in the instruction and tested by the verifier; removed requirements must disappear from both the oracle and tests. The container may expose task fixtures but may not package the oracle solution or hidden tests into the agent-visible environment.

Verification has two layers. Static checks reject malformed schemas, unsafe paths, missing required files, inconsistent reward output, excessive prompt length, and overlap with held-out evaluation tasks. Executable checks then build the isolated environment, run the oracle solution, and invoke the generated tests. A task is accepted only when the verifier emits reward 1. This gold reward certifies internal task consistency; it is distinct from the downstream policy reward used to estimate learning difficulty.

\subsection{Preprocessing, Training, and Evaluation Protocol}
\label{sec:app_protocol}

\paragraph{Normalization and filtering.}
Accepted bundles are normalized into a common Terminal-Bench-style schema before conversion to training records. Each method contributes exactly 100 accepted bundles to the downstream RL source; failed candidates, repair attempts, and intermediate materializations are excluded from that source. The conversion verifies the Docker specification and test entry point, standardizes task metadata, and drops invalid filesystem artifacts. Prompt length is computed with the target tokenizer, the actual agent system prompt, and the full chat template. The experimental protocol uses a 4096-token threshold for the smaller settings and an 8192-token threshold for the 35B setting; filtering statistics are recorded before training, and prompts are not silently truncated.

\paragraph{Preflight validation.}
Before a training run, the pipeline loads every normalized task, rechecks the configured train/evaluation split and overlap filters, builds representative containers, and executes the same oracle-plus-test path used during synthesis. These checks are completed before model workers are launched so that container or verifier failures cannot consume rollout budget. Seed and evaluation splits remain separate throughout selection and downstream evaluation.

\paragraph{Optimization and rollout configuration.}
The reported 35B policy is trained with GRPO using test-derived rewards and no learned reward model. Training uses FSDP2 with parameter and activation offload, gradient checkpointing, and bfloat16 computation. vLLM generates asynchronous rollouts with eight samples per prompt, temperature 1.0, top-$p$ 0.9, a 1024-token response cap, and at most 50 agent steps; the key-value cache uses FP8. The 35B run uses two H800 80\,GB GPUs for training, while Pass@1 evaluation uses one H800 80\,GB GPU. All methods use the same downstream training and evaluation protocol; only their synthesized training sources differ.

\paragraph{Evaluation scope.}
The evaluation compares the complete frontier-aware and verifier-backed pipeline under the shared downstream protocol. Each method exports 100 verified environments, and its synthesis totals remain within the same broad operational range as the other methods. A solver-off comparison over an identical candidate pool would add a component-level analysis of selector behavior.

\subsection{Qualitative Frontier Cases}
\label{sec:app_cases}

\begin{figure*}[p]
  \centering
  \small
  \begin{tcolorbox}[
    colback=red!2!white,
    colframe=red!75!black,
    colbacktitle=red!75!black,
    coltitle=white,
    title={Increase complexity (\texttt{in\_depth}): harden easy and near-frontier seeds},
    fonttitle=\bfseries\small,
    boxrule=0.35mm,
    width=\textwidth,
    arc=1.5mm,
    auto outer arc,
    boxsep=0mm,
    left=1.5mm,
    right=1.5mm,
    top=1mm,
    bottom=1mm
  ]
    \centering
    \begin{tikzpicture}[
      arrow/.style={-{Latex[length=1.8mm]}, line width=0.4pt, draw=red!70!black},
      bench/.style={font=\bfseries\scriptsize, align=right, text width=0.10\textwidth, inner sep=2pt},
      seed/.style={draw=gray!60!black, fill=gray!8, rounded corners=1mm, align=left, text width=0.22\textwidth, inner sep=3pt, font=\scriptsize},
      action/.style={draw=red!65!black, fill=red!5, rounded corners=1mm, align=left, text width=0.30\textwidth, inner sep=3pt, font=\scriptsize},
      take/.style={draw=red!45!black, fill=white, rounded corners=1mm, align=left, text width=0.245\textwidth, inner sep=3pt, font=\scriptsize}
    ]
      \matrix[column sep=2.2mm, row sep=2.8mm] {
        \node[bench] (bashlabel) {Case I\par Bash logs\par \textit{near frontier}}; &
        \node[seed] (bashseed) {\textbf{Seed contract}\par Recursively process nested \texttt{.log} files; update processed logs and per-directory summaries; remain idempotent under repeated and concurrent runs.\par \textbf{Capability:} $\hat p=0.747$ (slightly easy).}; &
        \node[action] (bashaction) {\textbf{MILP: \texttt{increase\_complexity:in\_depth}}\par Frontier score $0.9999$; projected $\hat p'=0.497$.\par \textbf{Materialization: JSON log processor}\par Adds structured summaries, special-character filenames, C-locale sorting, file locks, and atomic writes.}; &
        \node[take] (bashtake) {\textbf{Skill and artifact effect}\par Preserves Bash traversal and concurrency while strengthening deterministic structured output. Instruction, fixtures, oracle, tests, and container are co-edited.\par \textbf{Verifier:} static pass; oracle reward $1.0$.}; \\
        \node[bench] (mergelabel) {Case II\par Data merger\par \textit{too easy}}; &
        \node[seed] (mergeseed) {\textbf{Seed contract}\par Merge CSV, JSONL, and JSON records into one CSV with email, phone, and status fields.\par \textbf{Capability:} $\hat p=1.000$ (saturated).}; &
        \node[action] (mergeaction) {\textbf{MILP: \texttt{increase\_complexity:in\_depth}}\par Frontier score $0.4578$; projected $\hat p'=0.750$.\par \textbf{Materialization: deterministic merger}\par Requires an exact email union, empty-string missing values, standards-compliant quoting, exact headers, and sorted rows.}; &
        \node[take] (mergetake) {\textbf{Skill and artifact effect}\par Preserves multi-source parsing but makes edge cases and output order executable test conditions rather than prompt-only wording.\par \textbf{Verifier:} static pass; oracle reward $1.0$.}; \\
      };
      \draw[arrow] (bashseed) -- (bashaction);
      \draw[arrow] (bashaction) -- (bashtake);
      \draw[arrow] (mergeseed) -- (mergeaction);
      \draw[arrow] (mergeaction) -- (mergetake);
    \end{tikzpicture}
  \end{tcolorbox}

  \vspace{1mm}
  \begin{tcolorbox}[
    colback=blue!2!white,
    colframe=blue!70!black,
    colbacktitle=blue!70!black,
    coltitle=white,
    title={Reduce complexity (\texttt{in\_depth}): construct deterministic bridge environments},
    fonttitle=\bfseries\small,
    boxrule=0.35mm,
    width=\textwidth,
    arc=1.5mm,
    auto outer arc,
    boxsep=0mm,
    left=1.5mm,
    right=1.5mm,
    top=1mm,
    bottom=1mm
  ]
    \centering
    \begin{tikzpicture}[
      arrow/.style={-{Latex[length=1.8mm]}, line width=0.4pt, draw=blue!65!black},
      bench/.style={font=\bfseries\scriptsize, align=right, text width=0.10\textwidth, inner sep=2pt},
      seed/.style={draw=gray!60!black, fill=gray!8, rounded corners=1mm, align=left, text width=0.22\textwidth, inner sep=3pt, font=\scriptsize},
      action/.style={draw=blue!60!black, fill=blue!5, rounded corners=1mm, align=left, text width=0.30\textwidth, inner sep=3pt, font=\scriptsize},
      take/.style={draw=blue!40!black, fill=white, rounded corners=1mm, align=left, text width=0.245\textwidth, inner sep=3pt, font=\scriptsize}
    ]
      \matrix[column sep=2.2mm, row sep=2.8mm] {
        \node[bench] (systemdlabel) {Case III\par Systemd logs\par \textit{too hard}}; &
        \node[seed] (systemdseed) {\textbf{Seed contract}\par Deploy a systemd unit, JSON-logging application, rsyslog, logrotate, and journal-based monitoring.\par \textbf{Capability:} $\hat p=0.000$ (unsolved in recorded rollouts).}; &
        \node[action] (systemdaction) {\textbf{MILP: \texttt{reduce\_complexity:in\_depth}}\par Frontier score $0.4578$; projected $\hat p'=0.250$.\par \textbf{Materialization: log-analyzer bridge}\par Replaces live services with JSON fixtures; detects restart loops, counts warnings/errors, ranks recent errors, and flags malformed fields.}; &
        \node[take] (systemdtake) {\textbf{Skill and artifact effect}\par Preserves log parsing, validation, failure detection, ranking, and structured reports while removing service deployment and concurrency noise.\par \textbf{Verifier:} static pass; oracle reward $1.0$.}; \\
        \node[bench] (tokenlabel) {Case IV\par Token service\par \textit{too hard}}; &
        \node[seed] (tokenseed) {\textbf{Seed contract}\par Implement Spring Boot access/refresh rotation, one-time WebSocket tokens, database state, and race-safe concurrent operations.\par \textbf{Capability:} $\hat p=0.000$.}; &
        \node[action] (tokenaction) {\textbf{MILP: \texttt{reduce\_complexity:in\_depth}}\par Frontier score $0.4578$; projected $\hat p'=0.250$.\par \textbf{Materialization: file-level validator}\par Uses deterministic configuration, token events, and reference time; requires validation, repair, status labeling, risk scoring, and ranking.}; &
        \node[take] (tokentake) {\textbf{Skill and artifact effect}\par Preserves expiry, revocation, token consumption, repair, and security-state reasoning while removing the web stack, database, and races.\par \textbf{Verifier:} static pass; oracle reward $1.0$.}; \\
      };
      \draw[arrow] (systemdseed) -- (systemdaction);
      \draw[arrow] (systemdaction) -- (systemdtake);
      \draw[arrow] (tokenseed) -- (tokenaction);
      \draw[arrow] (tokenaction) -- (tokentake);
    \end{tikzpicture}
  \end{tcolorbox}

  \vspace{1mm}
  \begin{tcolorbox}[
    colback=green!2!white,
    colframe=green!50!black,
    colbacktitle=green!50!black,
    coltitle=white,
    title={Diversify frontier (\texttt{in\_breadth}): vary the instance without shifting the skill family},
    fonttitle=\bfseries\small,
    boxrule=0.35mm,
    width=\textwidth,
    arc=1.5mm,
    auto outer arc,
    boxsep=0mm,
    left=1.5mm,
    right=1.5mm,
    top=1mm,
    bottom=1mm
  ]
    \centering
    \begin{tikzpicture}[
      arrow/.style={-{Latex[length=1.8mm]}, line width=0.4pt, draw=green!45!black},
      bench/.style={font=\bfseries\scriptsize, align=right, text width=0.10\textwidth, inner sep=2pt},
      seed/.style={draw=gray!60!black, fill=gray!8, rounded corners=1mm, align=left, text width=0.22\textwidth, inner sep=3pt, font=\scriptsize},
      action/.style={draw=green!45!black, fill=green!5, rounded corners=1mm, align=left, text width=0.30\textwidth, inner sep=3pt, font=\scriptsize},
      take/.style={draw=green!35!black, fill=white, rounded corners=1mm, align=left, text width=0.245\textwidth, inner sep=3pt, font=\scriptsize}
    ]
      \matrix[column sep=2.2mm] {
        \node[bench] (pgnlabel) {Case V\par PGN repair\par \textit{at frontier}}; &
        \node[seed] (pgnseed) {\textbf{Seed contract}\par Repair corrupted PGN games, validate legal chess moves, preserve headers, and emit structured analysis.\par \textbf{Capability:} $\hat p=0.533$ (near target).}; &
        \node[action] (pgnaction) {\textbf{MILP: \texttt{diversify\_frontier:in\_breadth}}\par Frontier score $0.9862$; projected $\hat p'=0.533$.\par \textbf{Materialization: diversified PGN repair}\par Changes the fixture and illegal move (\texttt{15. Kf9} $\rightarrow$ \texttt{15. Kf1}) while retaining legal-move checks and structured outputs.}; &
        \node[take] (pgntake) {\textbf{Skill and artifact effect}\par Preserves chess parsing, header handling, and validator structure while creating a neighboring instance at the same intended difficulty.\par \textbf{Verifier:} static pass; oracle reward $1.0$.}; \\
      };
      \draw[arrow] (pgnseed) -- (pgnaction);
      \draw[arrow] (pgnaction) -- (pgntake);
    \end{tikzpicture}
  \end{tcolorbox}

  \caption{Detailed agentic case study of policy-relative environment synthesis. Each row follows a seed contract through the MILP decision and synchronized materialization to its preserved learning signal and verification result. The capability estimate $\hat p$, projected rate $\hat p'$, and frontier score motivate each action: red panels denote complexity increases for easy seeds, blue panels denote deterministic reductions for seeds unsolved in the recorded rollouts, and the green panel denotes frontier-preserving diversification. All five materializations co-edit the required task artifacts, pass static validation, and obtain oracle reward $1.0$. These traces illustrate selection and consistency semantics; downstream effects are evaluated by the benchmark results.}
  \label{fig:agentic_case_study}
\end{figure*}
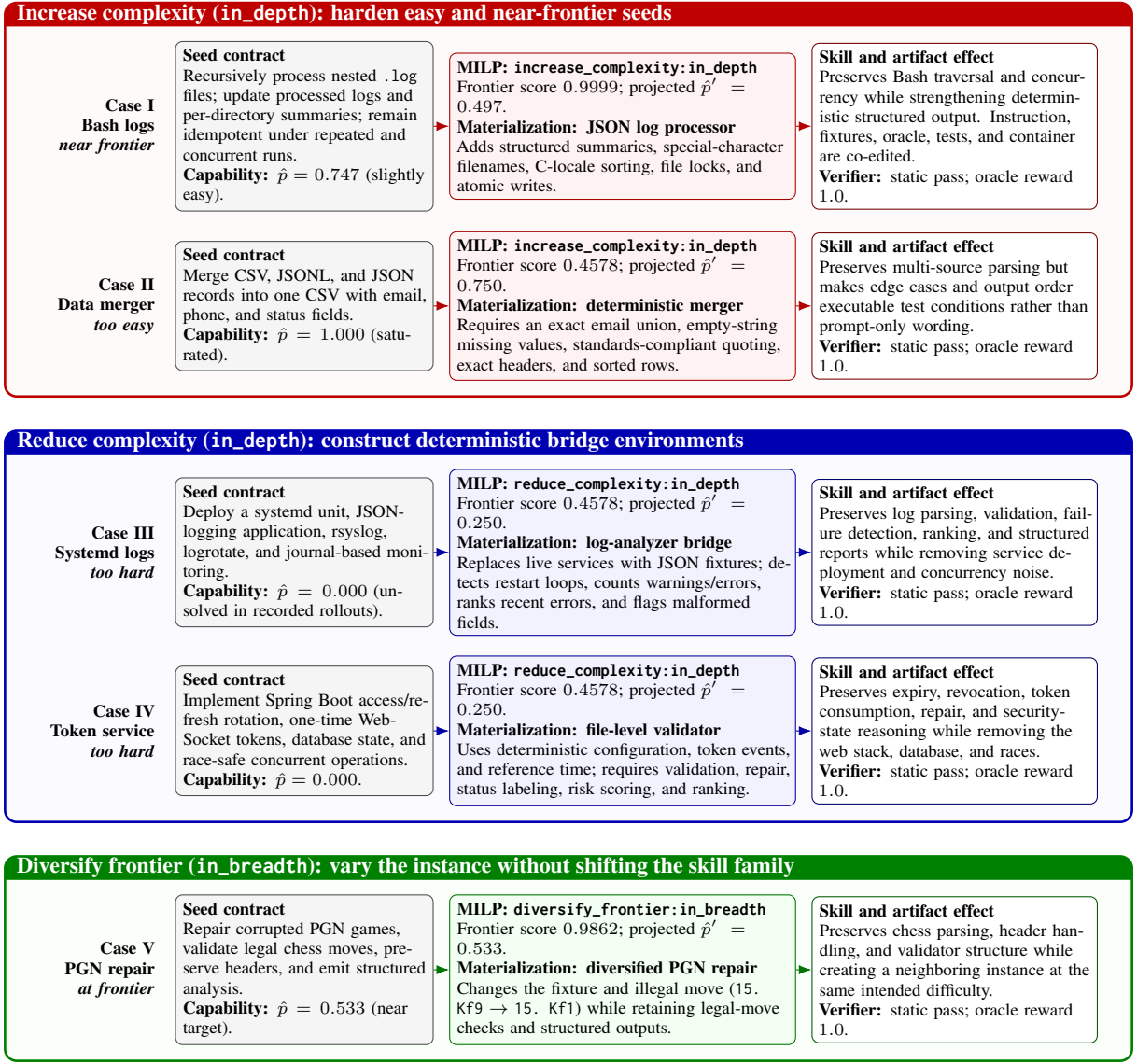

Figure~\ref{fig:agentic_case_study} traces five verified environments from seed contracts through policy-relative action selection, synchronized materialization, and executable verification. For each case, the figure reports the estimated seed pass rate, the projected rate and frontier score under the selected action, the concrete contract changes, and the preserved skill path. The red, blue, and green panels cover complexity increase, complexity reduction, and frontier diversification. Original-to-synthesized instruction excerpts for all five cases appear in Section~\ref{sec:app_instruction_excerpts}.

\paragraph{Case I: frontier-guided hardening of a near-frontier Bash task.}
The Bash seed already exercises recursive traversal, concurrency, and idempotence, and its estimated pass rate of $0.747$ places it slightly on the easy side of the target frontier. The selected depth increase projects the pass rate to $0.497$ and receives a frontier score of $0.9999$. The materialized environment keeps the Bash execution path but replaces loosely structured text artifacts with deterministic JSON summaries, including nested traversal, unusual filenames, C-locale ordering, file locking, and atomic writes. These additions create observable failure modes in the generated tests instead of merely lengthening the instruction.

\paragraph{Case II: verifier-facing hardening of a saturated merger.}
The original multi-format merger has an estimated pass rate of $1.0$, so another close paraphrase would provide little learning signal. Its selected depth increase retains CSV, JSONL, and JSON parsing while projecting the rate to $0.750$. The new contract defines the exact union key, treatment of missing fields, standards-compliant quoting, header schema, and deterministic row order. A superficially plausible merge is insufficient: the agent must satisfy edge-case and artifact-level requirements that the oracle and tests enforce jointly.

\paragraph{Case III: a deterministic bridge from system deployment to log reasoning.}
The systemd seed combines application code with service installation, rsyslog, log rotation, and journal monitoring, yielding an estimated pass rate of $0.0$. Envs-FORGE selects reduction because the seed was unsolved in the recorded rollouts. The bridge replaces live services with deterministic JSON fixtures but preserves restart-loop detection, severity counting, malformed-record validation, recent-error ranking, and structured reporting. The projected rate rises to $0.250$, making the environment easier without discarding the log-analysis skill chain that motivated the seed.

\paragraph{Case IV: isolating security-state reasoning from a full web stack.}
The token-service seed is also estimated at $0.0$ because it couples refresh rotation and one-time WebSocket tokens with framework setup, persistent database state, and race-safe concurrency. The reduced environment fixes configuration, event records, and reference time, then asks the agent to validate, repair, label, score, and rank token states. This removes Java, web-server, database, and concurrency overhead while retaining expiry, revocation, consumption, repair, and risk reasoning. The result is a projected $0.250$ bridge task whose reward focuses on the intended security transitions rather than deployment failures.

\paragraph{Case V: frontier-preserving chess diversification.}
The PGN seed is already near the frontier at $0.533$, so increasing or reducing its intended difficulty is unnecessary. The selected breadth action retains chess parsing, header preservation, legal-move validation, and structured repair, but changes the corrupted game and concrete illegal move from \texttt{15. Kf9} to its legal repair \texttt{15. Kf1}. Its projected pass rate remains $0.533$ and the frontier score is $0.9862$. This example separates useful instance diversity from uncontrolled task drift: the fixture changes, whereas the skill family and verifier structure remain stable.

\paragraph{Cross-case interpretation.}
Together, the cases show why a mixed seed pool needs more than one fixed evolution direction. The same action space calls for hardening when a seed is saturated, reduction when infrastructure masks the target reasoning skill, and breadth diversification when the seed is already near the frontier. In every case, the instruction, fixtures, oracle, tests, and environment are updated under one artifact contract, and the oracle obtains reward 1 after static validation. These traces establish internal executability and synchronization; the downstream performance evidence is reported in Section~\ref{sec:main_results}.

\section{MILP Solution Procedure}
\label{sec:app_milp_solution}

This section details how the environment-synthesis policy in Section~\ref{sec:method_milp} is instantiated and solved. \method{} uses PySCIPOpt as the modeling interface and SCIP as the solver. The optimization unit is a synthesis action, not a generated token, artifact, or training step. For each seed, the candidate set combines one of three difficulty projections with an in-depth or in-breadth evolution direction. Each candidate specifies a complete pre-materialization contract: the intended difficulty movement, evolution direction, frontier score, and required or optional skill nodes. The MILP selects which contract is materialized into an executable environment.

\paragraph{Model assembly.}
For every eligible candidate $c=(i,a,d)$, the implementation creates a binary action variable $x_c$. In the core per-task instance, the local cardinality constraint selects exactly one of the six candidates for the current seed. The indexed equations also include a cardinality target $N$ for stacking local instances and allow shared skill-coverage constraints. Binary variables $u_{c,v}$ expose the skill subgraph induced by a selected action, and continuous variables $\xi_v$ record shortfall against configured coverage targets. Active slack is bounded by \(\bar{\xi}=0.2\) in the reported experiments. All frontier scores, eligibility indicators, and skill sets are computed before model construction, so Eqs.~\ref{eq:milp_objective}--\ref{eq:slack_cap} remain linear. Candidates that fail deterministic eligibility, overlap, or prompt-length checks are removed before optimization.

The formulation requires no big-$M$ constants. A required skill satisfies $u_{c,v}=x_c$, a forbidden skill is fixed to zero, and an optional skill satisfies $u_{c,v}\leq x_c$ with a small activation penalty. The implementation allocates these linking variables over a uniform candidate--skill index to keep the decoded solver payload auditable. The primary combinatorial choice is which of the six synthesis-action variables to activate for each seed. Skill-variable values are fixed or bounded by $x_c$, and SCIP presolve can remove much of this deterministic structure. Soft coverage terms use \(\xi_v\) to tolerate at most 0.2 shortfall on active targets.

\paragraph{Branch-and-cut search.}
For this linear mixed-integer model, SCIP's exact search is appropriately described as \emph{branch-and-cut}. Presolve fixes linked skill variables, removes redundant rows, and tightens the remaining domains. SCIP then solves LP relaxations to obtain dual bounds, branches on fractional action decisions, separates valid cutting planes, propagates bound changes through the constraint system, and applies primal heuristics to obtain feasible action sets. Search continues until the primal--dual gap is closed and optimality is certified, or until SCIP returns another termination status. The implementation reports that status explicitly before any decoded action is used.

\paragraph{Recorded environment-synthesis solve.}
The recorded run contains 100 seeds and six projection--direction candidates per seed, giving 600 primary action variables. It requests 100 synthesized environments, uses active coverage slack cap \(\bar{\xi}=0.2\), and PySCIPOpt reports \texttt{status=optimal} with objective value $49.9104$. Since the target count equals the seed count, the run still exports one selected action per seed. The coverage slack records bounded shortfall for active skill targets and conditions the decoded skill payload; it does not change the downstream export size of 100 verified environments. Alternative portfolio settings with smaller \(N\) or tighter coverage slack would create a stronger cross-seed allocation problem.

\paragraph{Solution decoding and audit trail.}
The implementation accepts the SCIP path only when PySCIPOpt reports \texttt{optimal}. Selected $x$ and $u$ variables are decoded with a $0.5$ threshold, after which the action contract and induced skill subgraph are passed to the environment-synthesis prompt. The trace records the backend, formulation, status, objective value, candidate and target counts, requested coverage, realized slack, selected actions, and selected skill nodes. The objective value is a solver audit quantity for Eq.~\ref{eq:milp_objective}; downstream Pass@1 is measured separately after RL training.

\paragraph{Fallback and correctness boundary.}
For portability, an unavailable PySCIPOpt installation or a non-optimal SCIP status triggers a backend-explicit fallback: exact enumeration is attempted only when the candidate list contains at most 24 entries, and larger instances use a deterministic coverage-aware greedy policy. Every selected item retains the backend label, and the recorded planning trace uses PySCIPOpt rather than either fallback. Solver optimality certifies the discrete selection problem defined by the input coefficients and constraints. Semantic consistency of the generated environment is certified separately by synchronized artifact materialization and the post-generation gold-solution verifier described in Section~\ref{sec:app_artifact_contract}.

\section{Prompt and Task-Instruction Excerpts}
\label{sec:app_prompts}

This section reproduces the prompt clauses that determine synthesis strategy and artifact consistency. For readability, we omit package-manager and mirror fallbacks, repeated JSON-schema boilerplate, task-specific file locations, and repeated output-format instructions. The excerpts retain the clauses that distinguish the compared strategies, define the synchronized-artifact contract, and specify task semantics.

\tcbset{
  envspromptbox/.style={
    enhanced,
    breakable,
    colback=gray!3!white,
    colframe=gray!60!black,
    colbacktitle=gray!70!black,
    coltitle=white,
    fonttitle=\bfseries\small,
    fontupper=\footnotesize,
    boxrule=0.35mm,
    arc=1mm,
    auto outer arc,
    left=1.5mm,
    right=1.5mm,
    top=1mm,
    bottom=1mm,
    before skip=4pt,
    after skip=6pt
  }
}

\subsection{Shared Baseline Synthesis Contract}
\label{sec:app_baseline_prompts}

The four prompting baselines call the same artifact-generation template. They differ only in the strategy and evolution-direction fields injected after the shared contract.

\begin{tcolorbox}[
  envspromptbox,
  colframe=blue!60!black,
  colbacktitle=blue!65!black,
  title={Shared baseline prompt (core excerpt)}
]
{\ttfamily\raggedright Synthesize one Terminal-Bench-style task directory from the seed task. Return exactly one JSON object, with top-level keys task\_name, instruction\_md, files, and metadata.\par}

\begin{itemize}[leftmargin=1.3em,itemsep=1pt,topsep=3pt]
  \item Create a small but real task variant rather than restating the seed.
  \item Rewrite the user-facing instruction and synchronize the oracle solution, test runner, verifier tests, execution environment, and any changed fixtures or helper files.
  \item Keep the container build context valid, pin the runtime, and ensure that every command used by the solution and tests exists in the generated environment.
  \item Make the tests verify the generated instruction and solution. Every tested threshold, tie-break, filename, output key, or fixture value must be stated in the instruction rather than hidden in the verifier.
  \item Make the test runner write reward 1 on success and reward 0 on failure, including when the underlying test command fails.
  \item Before returning, mentally execute the build, oracle solution, and verifier; the expected result is reward 1.
\end{itemize}

The runtime suffix supplies the selected strategy instruction, evolution direction, sample index, synthesis model, seed name, and shortened seed files for structural reference.
\end{tcolorbox}

\begin{tcolorbox}[
  envspromptbox,
  breakable=false,
  colframe=blue!60!black,
  colbacktitle=blue!65!black,
  title={Baseline-specific strategy injection}
]
\noindent\textbf{\texttt{few\_shot}; direction: none.}
``Create a close variant that preserves the seed task structure while changing concrete requirements and verifier expectations.''

\smallskip
\noindent\textbf{\texttt{self\_instruct}; direction: none.}
``Invent a related terminal task in the same domain with its own task description, solution, and verifier.''

\smallskip
\noindent\textbf{\texttt{evol\_instruct\_in\_depth}; direction: \texttt{in\_depth}.}
``Deepen the seed by adding a harder constraint or edge case and update solution and tests accordingly.''

\smallskip
\noindent\textbf{\texttt{evol\_instruct\_in\_breadth}; direction: \texttt{in\_breadth}.}
``Broaden the seed into a neighboring task type while keeping the same Terminal-Bench directory contract.''
\end{tcolorbox}

The shared template controls artifact completeness, whereas the injected sentence fixes how the baseline moves from the seed. Few-shot stays structurally close, Self-Instruct permits a related same-domain task, and the two Evol-Instruct variants commit to depth or breadth before observing the policy-relative reward band.

\begin{tcolorbox}[
  envspromptbox,
  colframe=blue!60!black,
  colbacktitle=blue!65!black,
  title={Shared baseline repair prompt (core excerpt)}
]
{\ttfamily\raggedright Repair a failed generated Terminal-Bench-style task directory. Return one complete JSON object rather than patch fragments.\par}

\begin{itemize}[leftmargin=1.3em,itemsep=1pt,topsep=3pt]
  \item Make \texttt{instruction\_md} identical to the generated instruction file after trimming whitespace.
  \item Make the oracle solve exactly that instruction and make both verifier layers test exactly that instruction and oracle behavior.
  \item Include every changed fixture or helper file, preserve a buildable environment context, and pin the runtime.
  \item State exact verifier expectations in the instruction and always emit reward 0 or 1 even when the test command fails.
  \item Re-evaluate the build--solve--test path and target oracle reward 1.
\end{itemize}

The repair call receives the strategy, evolution direction, repair index, failure summary, previous artifact JSON, and seed files. Repair preserves the baseline policy while correcting parsing, build, static-validation, or oracle-verification failures.
\end{tcolorbox}

\subsection{Solver-Conditioned Envs-FORGE Prompt}
\label{sec:app_forge_prompt}

Envs-FORGE retains the same complete-environment output contract but conditions generation on an optimized action. The prompt receives a solver payload, a projection instruction, an evolution-direction instruction, an artifact edit mask, and a skill subgraph before it sees the seed files.

\begin{tcolorbox}[
  envspromptbox,
  breakable=false,
  colframe=green!45!black,
  colbacktitle=green!45!black,
  title={Solver payload and Envs-FORGE objective (core excerpt)}
]
The payload records the seed pass-rate estimate and reward band, the selected \texttt{projection}, \texttt{evolution\_direction}, and combined \texttt{action\_variant}, together with the projected pass rate, frontier score, and target difficulty. The prompt objective is:

\begin{itemize}[leftmargin=1.3em,itemsep=1pt,topsep=3pt]
  \item project the seed toward the current policy's learning frontier;
  \item preserve the target skill subgraph rather than jumping to an unrelated easy task;
  \item increase easy seeds, reduce hard seeds into bridge tasks, and diversify frontier seeds near their current difficulty;
  \item materialize the selected action as one synchronized executable environment.
\end{itemize}

The six possible action variants are the Cartesian product of \texttt{increase\_complexity}, \texttt{reduce\_complexity}, or \texttt{diversify\_frontier} with \texttt{in\_depth} or \texttt{in\_breadth}.
\end{tcolorbox}

\begin{tcolorbox}[
  envspromptbox,
  colframe=green!45!black,
  colbacktitle=green!45!black,
  title={Projection and evolution-direction instructions}
]
\noindent\textbf{\texttt{increase\_complexity}.}
The seed appears easy. Add one or two concrete constraints, edge cases, stricter output requirements, or a larger fixture while preserving the seed skill subgraph.

\smallskip
\noindent\textbf{\texttt{reduce\_complexity}.}
The seed appears beyond the current agent. Split away secondary systems or create a smaller bridge task while preserving the target skill subgraph and its important artifacts.

\smallskip
\noindent\textbf{\texttt{diversify\_frontier}.}
The seed is near the frontier. Keep difficulty similar, diversify data and constraints, and avoid large simplification or unrelated breadth jumps.

\smallskip
\noindent\textbf{\texttt{in\_depth}.}
Strengthen constraints and edge cases without changing the core skill graph.

\smallskip
\noindent\textbf{\texttt{in\_breadth}.}
Vary fixtures, neighboring constraints, or surrounding context while keeping difficulty in the same neighborhood.
\end{tcolorbox}

\begin{tcolorbox}[
  envspromptbox,
  colframe=green!45!black,
  colbacktitle=green!45!black,
  title={Artifact edit masks (core operators)}
]
Let \textsc{Inst}, \textsc{Data}, \textsc{Sol}, \textsc{Test}, \textsc{Env}, and \textsc{Help} denote the instruction, fixtures, oracle solution, tests, environment, and helper files.

\noindent\textbf{\texttt{increase\_complexity}.}
Required: \textsc{Inst}, \textsc{Data}, \textsc{Sol}, \textsc{Test}; optional: \textsc{Env}, \textsc{Help}. Operators add explicit constraints and edge cases, change or enlarge fixtures, update the algorithm, and strengthen checks for boundary cases and deterministic outputs.

\smallskip
\noindent\textbf{\texttt{reduce\_complexity}.}
Required: \textsc{Inst}, \textsc{Data}, \textsc{Sol}, \textsc{Test}, \textsc{Env}; optional: \textsc{Help}. Operators split secondary systems, replace live dependencies with deterministic fixtures, rewrite the solution for a bridge task, test the core skill only, and simplify the runtime.

\smallskip
\noindent\textbf{\texttt{diversify\_frontier}.}
Required: \textsc{Inst}, \textsc{Data}, \textsc{Sol}, \textsc{Test}; optional: \textsc{Env}, \textsc{Help}. Operators reframe at similar difficulty, replace fixtures while preserving scale, update expected logic while retaining the solution shape, and preserve verifier granularity.
\end{tcolorbox}

\begin{tcolorbox}[
  envspromptbox,
  colframe=green!45!black,
  colbacktitle=green!45!black,
  title={Skill-subgraph and generation contract}
]
The skill-subgraph prompt exposes three sets. \texttt{preserve\_nodes} must remain observable in the generated instruction and verifier; additions must come from \texttt{allowed\_added\_nodes}; and a reduction may remove or simulate \texttt{allowed\_removed\_nodes} only when the preserved core remains testable. The model must treat every artifact in \texttt{required\_artifacts} as a mandatory synchronized edit and restrict each edit to the operator families in the selected mask.

The remaining clauses require a compact runnable artifact, favor reuse of the seed runtime, permit deterministic file-level simulations for reduced system tasks, prohibit hidden test constraints, and require the same build--oracle--verifier mental execution used by the baselines. The returned metadata stores the solver payload alongside the target difficulty and projection summary.
\end{tcolorbox}

The difference from a fixed baseline appears upstream of generation: a baseline injects one predetermined strategy sentence, whereas Envs-FORGE injects a policy-relative solver decision plus explicit skill and artifact constraints. Both families are held to the same executable-environment and gold-verification standard.

\subsection{Original-to-Synthesized Instruction Excerpts}
\label{sec:app_instruction_excerpts}

The following five pairs retain the clauses that define the task, its edge cases, and its verifier-facing outputs. Repeated file locations and delivery boilerplate are omitted. The colors match Figure~\ref{fig:agentic_case_study}.

\begin{tcolorbox}[
  envspromptbox,
  breakable=false,
  colframe=red!70!black,
  colbacktitle=red!70!black,
  title={Case I: Bash log processor $\rightarrow$ JSON log processor}
]
\textbf{Original instruction, core clauses.}
\begin{itemize}[leftmargin=1.3em,itemsep=1pt,topsep=2pt]
  \item Fix a Bash script that recursively processes every log file, records each filename and line count in a global output and per-directory summaries, and sorts by filename.
  \item Make all updates atomic, restartable, deterministic, and safe under repeated or concurrent execution; process each file exactly once and handle spaces or special characters in filenames.
\end{itemize}

\textbf{Synthesized instruction, core clauses.}
\begin{itemize}[leftmargin=1.3em,itemsep=1pt,topsep=2pt]
  \item Replace plain-text outputs with a processed JSON array, one JSON summary per directory, and a top-level directory-count summary.
  \item Recursively handle nested and unusual filenames, correctly count empty and non-newline-terminated files, sort under the C locale, write through temporary-file replacement, protect updates with a file lock, and remain idempotent.
\end{itemize}
\end{tcolorbox}

\begin{tcolorbox}[
  envspromptbox,
  breakable=false,
  colframe=red!70!black,
  colbacktitle=red!70!black,
  title={Case II: multi-format merger $\rightarrow$ deterministic sorted merger}
]
\textbf{Original instruction, core clauses.}
\begin{itemize}[leftmargin=1.3em,itemsep=1pt,topsep=2pt]
  \item Read one CSV, one JSONL file, and one JSON file with potentially non-standard schemas; output email, phone, and status for every unique email in the union.
  \item Do not print complete inputs, and emit a standard three-column CSV with a header.
\end{itemize}

\textbf{Synthesized instruction, core clauses.}
\begin{itemize}[leftmargin=1.3em,itemsep=1pt,topsep=2pt]
  \item Preserve the union semantics but require deterministic ascending email order and the exact header \texttt{email,phone,status}.
  \item Represent missing values as empty strings and use standards-compliant CSV quoting for commas, quotes, and newlines; the output must remain deterministic across runs.
\end{itemize}
\end{tcolorbox}

\begin{tcolorbox}[
  envspromptbox,
  breakable=false,
  colframe=blue!65!black,
  colbacktitle=blue!65!black,
  title={Case III: systemd log monitoring $\rightarrow$ log-analyzer bridge}
]
\textbf{Original instruction, core clauses.}
\begin{itemize}[leftmargin=1.3em,itemsep=1pt,topsep=2pt]
  \item Deploy a continuously running JSON-logging application as a systemd service, route its journal through rsyslog, rotate the resulting files, and provide a health-monitoring script.
  \item Detect restart loops and errors over a recent time window, emit a dynamic JSON report, and validate service management, routing, permissions, rotation, and end-to-end integration.
\end{itemize}

\textbf{Synthesized instruction, core clauses.}
\begin{itemize}[leftmargin=1.3em,itemsep=1pt,topsep=2pt]
  \item Analyze deterministic newline-delimited JSON fixtures instead of a live service stack. Detect more than three restart events in a five-minute window, count warnings and errors, and rank the five most recent errors.
  \item Record malformed JSON or missing fields, handle empty and unreadable fixtures without crashing, and emit one deterministic structured report using only the standard library.
\end{itemize}
\end{tcolorbox}

\begin{tcolorbox}[
  envspromptbox,
  breakable=false,
  colframe=blue!65!black,
  colbacktitle=blue!65!black,
  title={Case IV: full-stack token service $\rightarrow$ file-level token validator}
]
\textbf{Original instruction, core clauses.}
\begin{itemize}[leftmargin=1.3em,itemsep=1pt,topsep=2pt]
  \item Implement a Java/Spring service backed by H2 that issues access, refresh, and one-time WebSocket tokens with different lifetimes and storage semantics.
  \item Rotate refresh tokens atomically, ensure exactly one success under concurrent renewal or consumption, revoke all user tokens on logout, and periodically delete expired, revoked, or consumed state.
\end{itemize}

\textbf{Synthesized instruction, core clauses.}
\begin{itemize}[leftmargin=1.3em,itemsep=1pt,topsep=2pt]
  \item Read deterministic configuration, token-event records, and a fixed reference time; compute expiry, revocation, and one-time-token consumption status.
  \item Repair eligible recently used expired tokens, assign risk scores for expired, revoked, consumed, and high-risk-user conditions, then sort by descending risk and ascending token identifier in a structured JSON output.
\end{itemize}
\end{tcolorbox}

\begin{tcolorbox}[
  envspromptbox,
  breakable=false,
  colframe=green!45!black,
  colbacktitle=green!45!black,
  title={Case V: PGN repair and puzzles $\rightarrow$ diversified PGN repair}
]
\textbf{Original instruction, core clauses.}
\begin{itemize}[leftmargin=1.3em,itemsep=1pt,topsep=2pt]
  \item Repair multiple corrupted PGN games by detecting illegal, missing, and ambiguous moves; identify critical positions and tactical patterns.
  \item Solve FEN-based puzzles, including mate-in-$N$ instances, and return repaired games, puzzle continuations, and aggregate analysis.
\end{itemize}

\textbf{Synthesized instruction, core clauses.}
\begin{itemize}[leftmargin=1.3em,itemsep=1pt,topsep=2pt]
  \item Repair one corrupted game containing exactly one illegal move, \texttt{15. Kf9}, whose required correction is \texttt{15. Kf1}.
  \item Validate every move with \texttt{python-chess}, preserve all PGN headers, and return the original PGN, repaired PGN, and a structured error record. No other error is present.
\end{itemize}
\end{tcolorbox}

All five synthesized instructions were materialized with synchronized fixtures, oracle solutions, tests, and environments. Static validation passed and each oracle obtained reward 1. As in Figure~\ref{fig:agentic_case_study}, this establishes internal consistency of the generated task bundles; downstream policy accuracy is reported in the benchmark results.

\section{Compute Resources}
\label{sec:app_broader_impact}

For each prompt, we sample 8 rollouts with temperature $1.0$ and top-$p=0.9$.
The training batch size, PPO mini-batch size, and per-GPU micro-batch size are all set to $1$.
Both the maximum prompt length and maximum response length are set to $8192$ tokens.
Each trajectory allows at most $50$ agent-environment interaction steps, with a trajectory timeout of $900$ seconds.
The actor learning rate is set to $1\times10^{-6}$.
We train for one epoch, enable automatic checkpoint resume, and save a checkpoint at every step; after training finishes, only the final actor model weights are retained.
Rewards are computed solely from Terminal-Bench test outcomes, with test reward weight $1$ and judge reward weight $0$.
We do not include a KL term in the reward.
Training is conducted on 2 H800 80GB GPUs.
The actor, reference policy, and rollout engine share the same visible GPUs.
We use FSDP2 for distributed training, with gradient checkpointing, activation offloading, and FSDP offload policy enabled to reduce GPU memory usage.
Rollouts are generated asynchronously by the vLLM hybrid engine with tensor parallel size $2$.
The model weights use bfloat16 precision, the KV cache uses FP8, and both the maximum model length and maximum number of batched tokens are set to $32k$.

The current runs require two H800 80\,GB GPUs with Fully Sharded Data Parallel (FSDP) and CPU offload for training, plus one H800 80\,GB GPU for evaluation. Environment synthesis also requires candidate generation, MILP selection, container builds, and executable verification. Larger seed pools or larger backbones will increase synthesis, training, and evaluation cost.

\end{document}